\documentclass[article]{elsarticle}

\usepackage{hyperref}
\usepackage{graphicx}
\usepackage{graphics}
\usepackage{algorithm}
\usepackage{setspace}
\usepackage{algpseudocode}
\usepackage{epsfig}
\usepackage{times}
\usepackage{url}
\usepackage{multirow}
\usepackage{amsmath}
\usepackage{amssymb}
\usepackage{subcaption}
\usepackage{mathtools}
\usepackage{url}
\usepackage[utf8]{inputenc}
\usepackage{ dsfont }
\usepackage{ mathrsfs }
\usepackage{mathtools}
\usepackage{longtable}

\usepackage{adjustbox}

\usepackage{algorithm}
\usepackage[margin=1.5in]{geometry}

\algblockdefx{MRepeat}{EndRepeat}{\textbf{repeat}}{}
\algnotext{EndRepeat}

\usepackage{xcolor}

\journal{Expert Systems with Applications}

\usepackage{numcompress}\biboptions{authoryear}

\begin{document}

\begin{frontmatter}

\title{Monitoring electrical systems data-network equipment by means of Fuzzy and Paraconsistent Annotated Logic}

\author[mymainaddress]{Hyghor Miranda Côrtes}
\address[mymainaddress]{Centro Universit\'ario FEI, S\~ao Bernardo do Campo, SP, Brazil (hyghorcortes@gmail.com)}

\author[mymainaddress,mymainaddress2]{Paulo Eduardo Santos}
\address[mymainaddress2]{School of Science and Engineering, Flinders University, Adelaide, Australia (paedusan@gmail.com)}

\author[mymainaddress3]{João Inácio da Silva Filho}
\address[mymainaddress3]{Universidade Santa Cecília, Santos, SP, Brazil (inacio@unisanta.br)}

\begin{abstract}
The constant increase in the amount and complexity of information obtained from IT data network elements, for its correct monitoring and management, is a reality. The same happens to data networks in electrical systems that provide effective supervision and control of substations and hydroelectric plants. Contributing to this fact is the growing number of installations and new environments monitored by such data networks and the constant evolution of the technologies involved. This situation potentially leads to incomplete and/or contradictory data, issues that must be addressed in order to maintain a good level of monitoring and, consequently, management of these systems. In this paper, a prototype of an expert system is developed to monitor the status of equipment of data networks in electrical systems, which deals with inconsistencies without trivialising the inferences. This is accomplished in the context of the remote control of hydroelectric plants and substations by a Regional Operation Centre (ROC). The expert system is developed with algorithms defined upon a combination of Fuzzy logic and Paraconsistent Annotated Logic with Annotation of Two Values (PAL2v) in order to analyse uncertain signals and generate the operating conditions ({\em faulty, normal, unstable or inconsistent / indeterminate}) of the equipment that are identified as important for the remote control of hydroelectric plants and substations. A prototype of this expert system was installed on a virtualised server with CLP500 software (from the EFACEC manufacturer) that was applied to investigate scenarios consisting of a Regional (Brazilian) Operation Centre, with a Generic Substation and a Generic Hydroelectric Plant, representing a remote control environment.
\end{abstract}


\end{frontmatter}


\section{Introduction}
\label{sec:introducao}
Innovative computing techniques are deployed each year in order to increase the efficiency and responsiveness of the user needs in data networks, enhancing it both in scale and complexity. The need to constantly evaluate the status of these networks in this context is imminent, driving the development of monitoring systems with a wide range of features/functionalities. Network management has become an indispensable task from the network's correct functioning, an issue that is essential to maintaining the quality demanded by the market \citep{Stallings:1998}.

The continuous growth in the number and diversity of computer-network equipment and, consequently, in the volume of information signals coming from these, has turned network management into a complex task. This also applies to the field of automation, where automation manufacturers, with the popularisation of ethernet networks, started to provide communication modules with Simple Network Management Protocol (SNMP). The Industrial Automation Open Network (IAONA) protocol specification reinforces this trend by publishing a Management Information Base (MIB) for SNMP categorisation of industrial applications \citep{fonseca2006}.

As a result, manufacturers now provide Management Information Bases (MIBs) for their products, which add specific Object Identifiers (OIDs) for monitoring automation devices. Some examples of data about OIDs found in automation equipment in the MIBs and in the IAONA are:

\begin{itemize}
\item Application information (name, {\em in operation});
\item Protocol data (Ethernet / IP, ModbusTCP);
\item Device name, serial number, vendor name, device version
firmware.
\end{itemize}

According to \citep{fonseca2006,Villela19,flores09,freitas14}, in contrast to the current fast development of MIBs for data network automation, these devices usually end up not monitored or monitored in isolation from the rest of the plant. In this context, concentrating the monitoring data on a centralised data centre becomes a suitable solution. Supervisory Control and Data Acquisition (SCADA) systems at the operation centres provide some of the desirable features for monitoring data network devices involved in the automation of critical processes. 

Specifically in electrical systems, the need for monitoring is often more pronounced because there are completely isolated facilities and devices, without any monitoring events, not even local. This is particularly relevant in the context of the Brazilian power grids. In addition, there is an increasing amount of data to be analysed that are related to fault classification of network equipment, where contradictory and inaccurate information often occur. Therefore, in addition to more detailed monitoring, it is necessary to use algorithms that address this information gap. The application of algorithms with non-classical logic is an alternative to alleviate this problem, as they allow for a more refined, explicit, treatment of information than black-box systems, or traditional logic-based systems.

In this paper, Fuzzy logic and a Paraconsistent Annotated Logic with Annotation of Two Values (PAL2v)  are combined to analyse and classify the operating conditions of equipment in data networks within two SCADA-monitored electrical installations of a Regional Operations Centre (ROC) in Brazil. In this analysis, information (variables) are collected from each equipment in the power grid using two distinct sources (used as evidences), the degree of contradiction between these variables is calculated and then, through a hybrid Fuzzy-paraconsistent inference tool (PAL2v lattice), the condition of the equipment operation ({\em normal, failure, unstable or undetermined / inconsistent}) is inferred.

The next section presents the background upon which this work was developed. The method for the analysis of data networks proposed in this paper is described in Section \ref{parafuzzy} and  the related tests are described in Section \ref{exp}. Sections \ref{res} and \ref{disc} present the results and discussions, Section \ref{related} presents the related works, and Section \ref{conclu} close this paper with conclusions and future works.

\section{Background}

This section presents some fundamental concepts of automation, remote control of data network equipment, monitoring of electrical systems (Section \ref{2.1}), and also about Fuzzy logic and the Paraconsistent Annotated Logic with Annotation of Two Values (PAL2v) used in this work (Section \ref{fp-intro}).

\subsection{Concepts of automation and monitoring systems for the remote control of data network equipment of electrical systems}\label{2.1}

This section describes some concepts of the electrical substation automation systems, hydroelectric plants and operation centres, including Regional Operation Centres (ROC) and System Operations Centre (SOC), according to \citep{Jardini97,Paredes02,PE2003} that are in agreement with the technical procedure for the Brazilian electric system (Sub-Module 2.7), prepared by the National Operator of the Brazilian Electric System (ONS).

A substation automation system aims to provide the means for its operation and maintenance. It is characterised by two hierarchical levels: the interface level, with the process and data acquisition, and the level of command and supervision, also called central system.

\begin{figure}[ht!]
    \centering
\begin{subfigure}[b]{1\textwidth}
    \centering
    \includegraphics[scale=0.074]{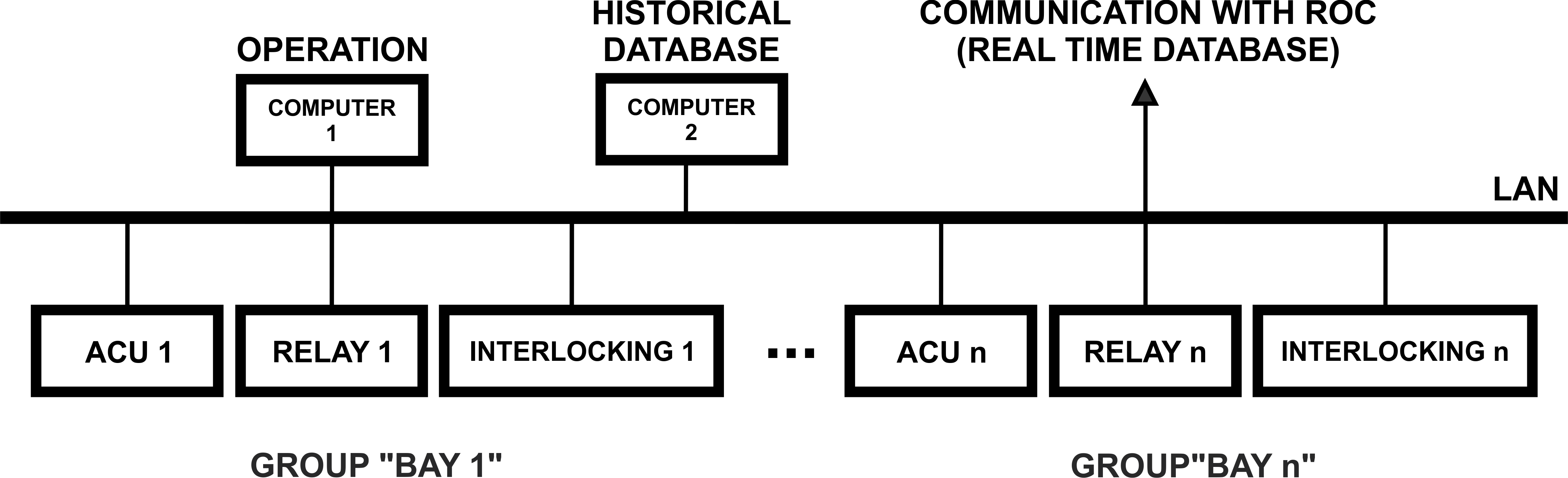}
    \caption{Substation}\label{1a}
\end{subfigure}
\hfill
\begin{subfigure}[b]{1\textwidth}
    \centering
    \includegraphics[scale=0.074]{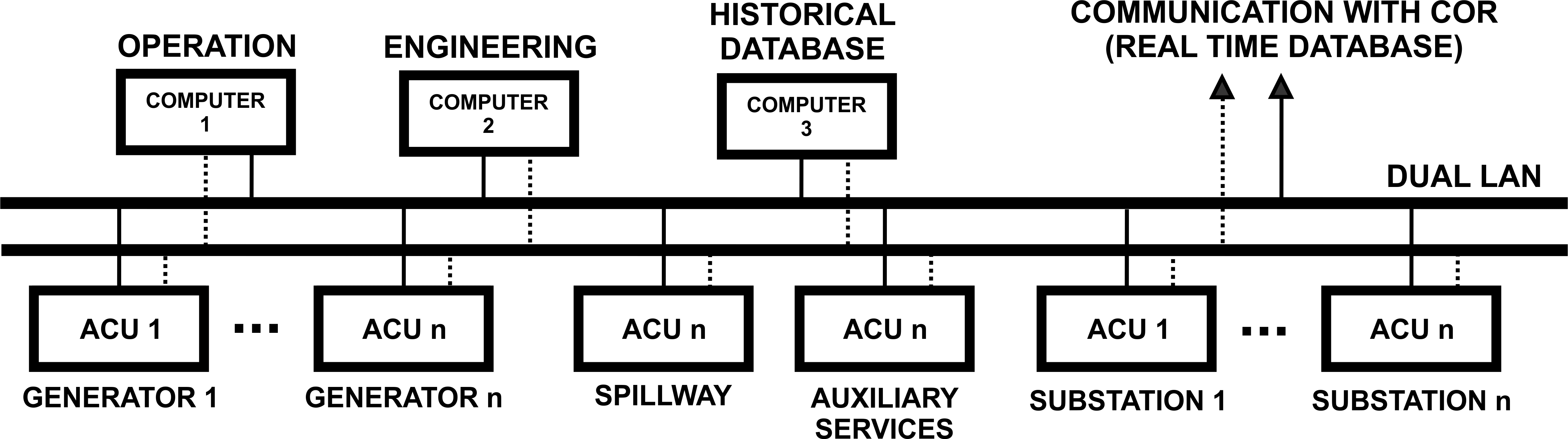}
    \caption{Hydroelectric plant}\label{1b}
\end{subfigure}
\hfill
\begin{subfigure}[b]{1\textwidth}
    \centering
    \includegraphics[scale=0.074]{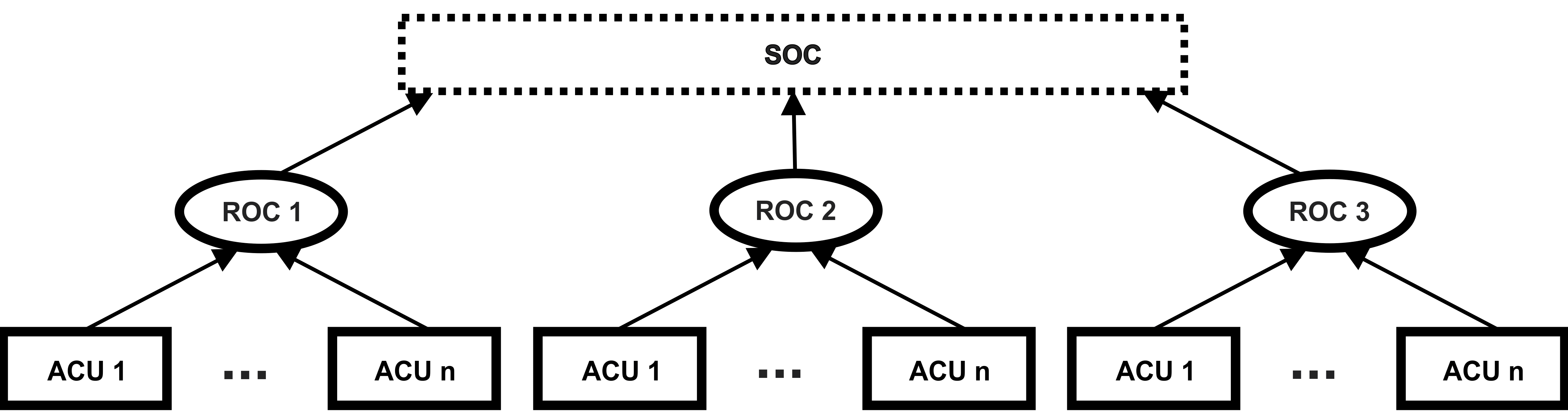}
    \caption{SOC, ROC, substations and hydroelectric plants}\label{1c}
\end{subfigure}
\caption{Typical architectures}\label{fig1}
\end{figure}

The interface level hosts the Acquisition and Control Units (ACUs), and some dedicated equipment such as protective relays, oscillography equipment, interlocking units, among others. The central system level is typically comprised of multiple workstations connected to a local area network that is connected to a process-level digital equipment. The central system has some functions such as monitoring the equipment status, measurement display, data monitoring protection, equipment control, alarm controls, and the indication of sequences of events. Figure \ref{1a} presents the typical architecture of a substation automation system, according to \citep{Jardini97,Paredes02,PE2003}.

The automation system of a hydroelectric plant also aims to provide the means for its operation and maintenance; it has a hierarchical-level configuration similar to a substation automation system, as shown in Figure \ref{1b}, and can display a dual local area network (LAN).
In this system, the process interface units can be composed of several modules, including a generator ACU, auxiliary services ACU, spillway ACU etc. These ACU units are integrated with speed regulators, voltage regulators, local controllers (in floodgates, pumps, compressors and others). Generally in large plants there are independent subsystems built for specific activities such as machine supervision (obtaining data about vibration, temperature, partial discharges etc.) and dam supervision.

In Regional Operation Centres (ROC) and System Operation Centre (SOC), substation and the hydroelectric plant automation are organised into a larger hierarchy of supervision and control subsystems:

\begin{itemize}

\item The Regional Operation Centre (ROC) concentrates the operation and service of substations and hydroelectric plants in a region. From the ROC, for example, remote control signals can be issued to start and stop a given generator in a plant, given the data collected from the ACUs. In short, ROC concentrates the Supervisory Control and Data Acquisition (SCADA) systems.

\item  The System Operation Centre (SOC) provides facilities for centralised system operation, generation and load coordination. SOC has a hardware and software structure with high-level functions, from which the information necessary for the proper and safe operation of the system is obtained. SOCs are often directly linked to ACUs, and a SCADA function is also included in each SOC.
\end{itemize}

The plant-automation system architecture can be represented as the hierarchy shown in Figure \ref{1c}.

 The most relevant points in a data network are usually monitored to ensure the reliability of SCADA supervision and the control of facilities. Some basic equipment, and related data, that are monitored are the following:

\begin{itemize}
\item ACUs (substations and hydroelectric plants): processing, memory, file allocation, protocol connections with ROCs SCADA systems and states of additional resources (communication cards, power supplies etc.).

\item Switches and routers: status of communication {\em ports} (traffic, latency, errors).

\item Servers (SOC / ROC): processing, memory, file allocation (disk units) data.

\end{itemize}

In this work, we considered the following sources of data: ACUs (information of protocol links between ACUs and SCADAs of the ROCs), routers (monitoring the outgoing traffic of substations and hydroelectric plants) and servers (providing data about the processing, memory and disk allocation).

\subsection{Fuzzy and Paraconsistent Annotated Logic}\label{fp-intro}

This section presents a brief introduction to Fuzzy set theory \citep{pedrycz1993fuzzy,ZADEH1965338}  and to the Paraconsistent Annotated Logic with Annotation of Two Values (PAL2v) \citep{dacosta1974,Abe06} used in this work.

\subsection*{Fuzzy Sets}

Let $U$ be an universe of discourse, a collection of objects $\{u\}$. A Fuzzy set $A\subset U$ is defined by a membership function $\mu_A$ that assumes values within the range [0,1]: \[\mu_A: U\rightarrow [0,1].\] 

The subset of  points $u\in U$ such that  $\mu_{A}\left ( u \right ) > 0$ is called the {\em support set of $A$}.

Let $A$ and $B$ be two distinct Fuzzy sets in $U$ with respective membership functions $\mu_{A}$ and $\mu_{B}$.  The Fuzzy set operations between $A$ and $B$ are defined as: 
\begin{itemize}
\item Union: $\mu_{A \cup B}(u)$ = $\mu_{A}(u)\; \delta\; \mu_{B}(u)$, where $\delta$ is a triangular co-norm (called t-conorm);
\item Intersection: $\mu_{A \cap  B}(u)$ = $\mu_{A}(u) \;\tau\; \mu_{B}(u)$, where $\tau$ is a triangular norm (called t-norm); and, 
\item Complement: $\mu_{\neg  A}(u)$ = 1 - $\mu_{A}(u)$.
\end{itemize}

A t-norm is a function $\tau:[0,1]\times [0,1]\rightarrow [0,1]$ that satisfies the properties of commutativity, monotonicity, associativity and has 1 as a neutral element. The following functions are t-norms,for instance: $TM(x,y)=min(x,y)$ (minimum or Gödel t-norm), $TP(x,y)=x.y$ (product t-norm)
$TL(x,y)=max(x+y-1,0)$ (Lukasiewicz t-norm) \citep{Navara:2007}.  A triangular co-norm ($t-conorm$) is a commutative, associative and monotonic function that has 0 as the neutral element. Examples of $t-conorm$s are: $SM(x,y)=max(x,y)$ (maximum or Gödel t-conorm), $SP(x,y)=x+y-x.y$ (product t-conorm, or probabilistic sum), $SL(x,y)=min(x+y,1)$ (Lukasiewicz t-conorm, or bounded sum) \citep{Navara:2007}. T-norms and t-conorms are generalisations of the classical conjunction and disjunction, respectively \citep{JAYARAM20092063}.

Given Fuzzy sets $A_{1}, A_{2}, ...,A_{n}$ defined respectively over the universes of discourse $U_{1}, U_{2}, ...,U_{n}$, an n-ary Fuzzy relation $R$ is a Fuzzy set in $U_{1}, U_{2}, ...,U_{n}$, expressed as: \[R = {[(u_{1}, u_{2}, ...,u_{n}), \mu _{R}(u_{1}, u_{2}, ...,u_{n})]|(u_{1}, u_{2}, ...,u_{n}) \in U_{1}, U_{2}, ...,U_{n}}\]

A Fuzzy knowledge base is represented by means of sets of Fuzzy conditional statements (Fuzzy rules). Let $A_{i}, B_{i}$ and $C_{i}$ be Fuzzy sets respectively defined over the universes of discourse $U, V$ and $W$. A Fuzzy rule such as: {\em if (x is $A_{i}$ and y is $B_{i}$) then (z is $C_{i}$)} can be interpreted as a the following Fuzzy $R_{i}$ relation:

\[\mu_{R_{i}} = [\mu_{A_i}(u)\;\; and\;\; \mu_{B_i}(v)]\rightarrow \mu_{C_i}(w),\;\; for \;u\in U, v \in V,\; w \in W,\]

where the symbol ``$\rightarrow$" is the Fuzzy implication. There are various types of Fuzzy implications \citep{JAYARAM20092063,4374117}. Assuming $x$ and $y$ such that $x, y\in [0,1]$, where $x$ is the antecedent and $y$ as the consequent of an implication ($x\rightarrow y$), the most usual interpretations of the Fuzzy implication  are the following \citep{RUAN199323}:
\begin{itemize}
    \item Zadeh implication \citep{zadeh75}: $\mathscr{I}(x,y) = max(1-x, min(x,y))$ 
    \item {\L}ukasiewicz implication \citep{zadeh75}: $\mathscr{I}(x,y) = min(1, 1 - x + y)$ 
    \item Mamdani implication \citep{1674779}: $\mathscr{I}(x,y) = min(x,y)$
\end{itemize}

In a Fuzzy system, each Fuzzy rule is represented by a Fuzzy relation. A set of Fuzzy relations describes the behaviour of the domain modelled by them. Therefore, a domain can be represented by a single Fuzzy relation, that is the combination of the set of relations describing it \citep{Gomide}. This combination can be obtained by applying the aggregation connective $aggreg$: {\em R = aggreg($R_{1}, R_{2}, ...,R_{n}$ )}. Usually, $aggreg$ is interpreted as set union ($max$ operator), however there are various possible aggregation connectives \citep{DUBOIS198585}.

In Fuzzy set theory, an important inference rule is the Generalised Modus Ponens (GMP) \citep{5408575,Cornelis2000} that is illustrated in the following example:
\begin{quote}
$x$ is $A$ \hspace{1.23in} (e.g., ``John is very tall"),\\
if ($x$ is $A$) then ($y$ is $B$)\hspace{0.3in} (e.g. ``If John is tall, then he is heavy")\\
$\therefore$ $y$ is $B$ \hspace{1.15in}(e.g. ``John is considerably heavy").
\end{quote}

Commonly this rule is interpreted by the law of compositional inference, suggested by \cite{5408575}. In this context, the rule {\em if $x$ is A then $y$ is B}, written as $A \rightarrow B$, is first transformed into a Fuzzy relation $R_{A \rightarrow B}$ that, assuming Mamdani's implication \citep{1674779}, becomes:
\[\mu_{A \rightarrow B} = min(\mu_A(u), \mu_B(v)); u\in U, v\in V\]

Given the fact that {\em if x is A} (or, simply $A$) and $R_{A \rightarrow B}$, Zadeh's compositional inference law states that:\[B = A \circ R_{A \rightarrow B},\]
where the composition of $A$ with $R_{A \rightarrow B}$ can be given by the max-min inference rule: \[\mu_{B}(v) = \underset{u}max({min(\mu_A(u), \mu_{R_{A \rightarrow B}}(u,v))})\]

\begin{figure}[ht!]
\centering
\includegraphics[scale=0.1]{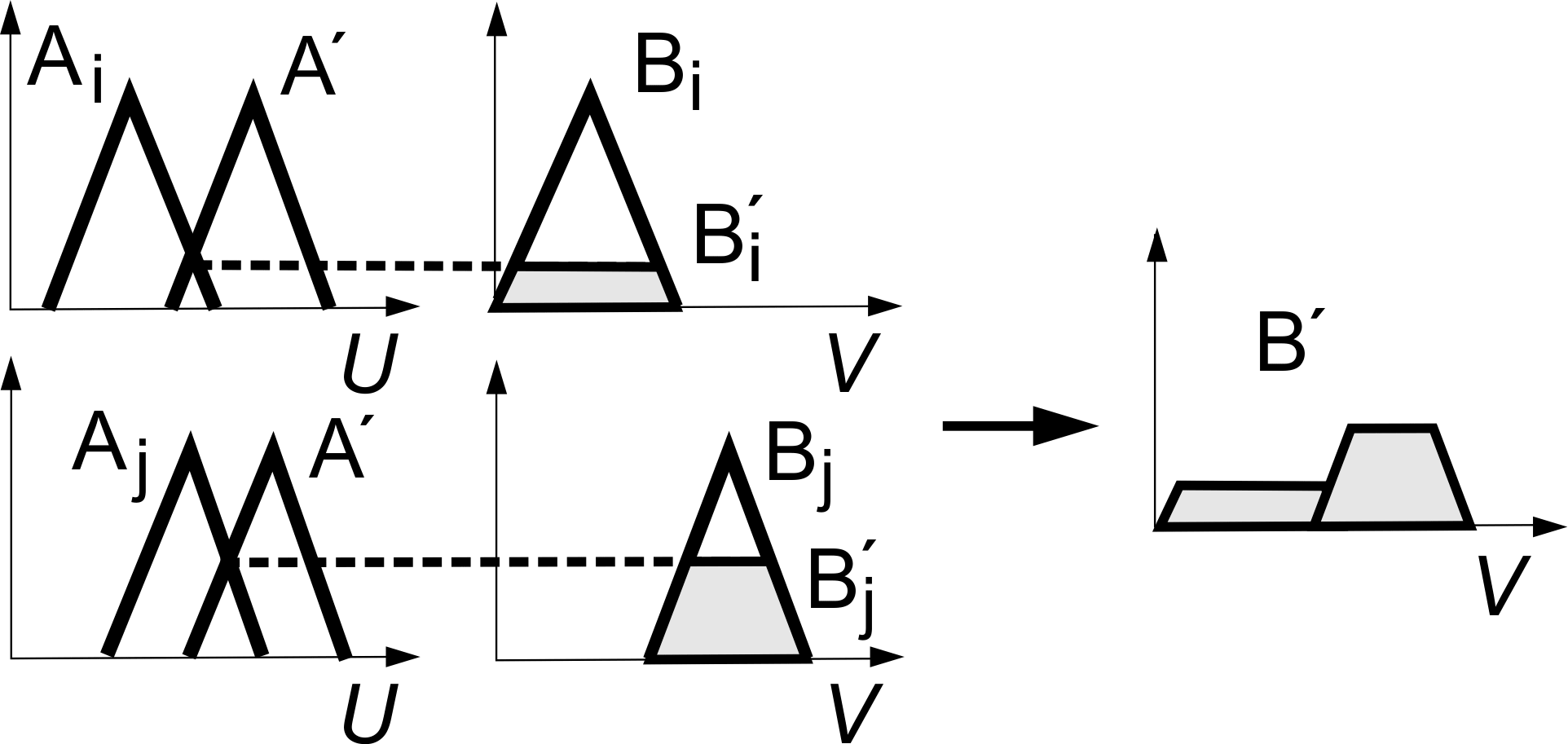}
\caption{Fuzzy inference engine \citep{Gomide}.}\label{2}
\end{figure}

The procedure above can be easily extended for any finite number of Fuzzy rules representing a domain. Figure \ref{2} \citep{Gomide} illustrates the $max-min$ inference process for two rules, $A_{i} \rightarrow B_{i}$ and $A_{j} \rightarrow B_{j}$, where $A'$ is the input fact and $B'$ is the output (the result of the Fuzzy inference procedure).

The Fuzzy modelling of a control system is based on translating the expert's knowledge into Fuzzy rules (as above), whereby the sensor data (which is the input to the system) is mapped to appropriate membership functions, this information is used to evaluate each of the Fuzzy rules representing the domain. The contributions of each of these rules is combined and then converted back into specific control output values. This conversion to control output value is called {\em defuzzification}, that extracts a {\em crisp} value from the results of the Fuzzy inference (for instance, extracting a single value form the Fuzzy set $B'$ in Figure \ref{2}). There are multiple defuzzification strategies available, and a description of them is outside the scope of this paper \citep{LEEKWIJCK1999159} .

 In the present work, this process of Fuzzy modelling of control systems is used to bridge sensor data from the electrical data network to logical states in a paraconsistent expert system.

\subsection*{Paraconsistent Annotated Logic}

Paraconsistent Logic (PL) \citep{dacosta1974,Abe15} is a non-classical logic that opposes the principle of non-contradiction and can be applied for the rigorous treatment of contradictory sensor information. \cite{Abe15ch} provides a clear description of the principles of Paraconsistent Logic. The present paper is based on a subset of PL, called Paraconsistent Annotated Logic.

Paraconsistent Annotated Logic (PAL) embraces concepts of uncertainty, inconsistency and incompleteness in its semantic structure, allowing reasoning with and about these concepts.

\begin{figure}[ht!]
\centering
\includegraphics[scale=0.10]{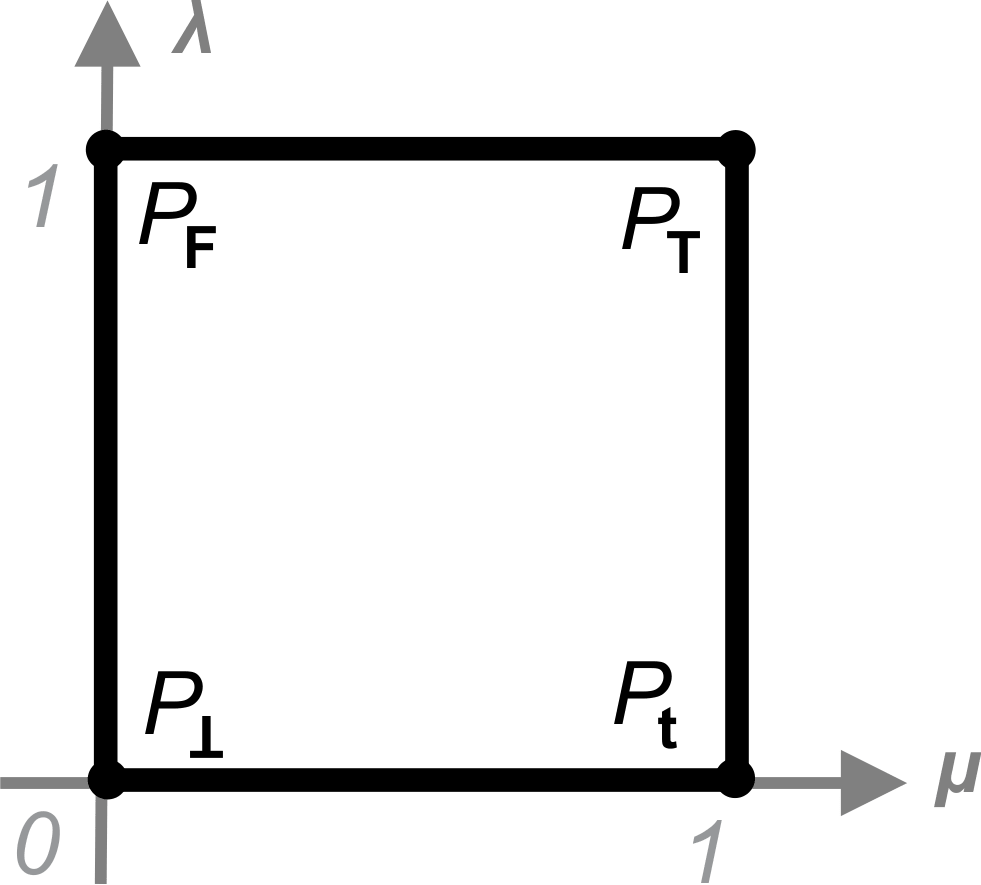}
\caption{PAL2v lattice.}\label{3}
\end{figure}

With this in mind, PAL has four truth values: $\tau$ = \{ $\top$ (inconsistency), t (truth), F (falsehood), and $\perp$ (paracomplete or indeterminate) \}. The set $\tau$ forms a complete lattice, characterised by the Hasse diagram shown in Figure \ref{3} \citep{Abe15ch}.

In this logic, the negation $\sim$ is considered in the following way: $\sim$(1)=0, $\sim$(0)=1, $\sim$($\top$)=$\top$ and $\sim$($\perp$)=$\perp$.

A proposition in PAL is accompanied by annotations ($\mu$), or degrees of evidence, that ascribe values (or {\em logical connotations} \citep{filho12}) to the proposition. For instance, given a proposition $P$ and a related annotation $\mu$, $P\mu $ can be interpreted as: “I believe the proposition $P$ with a maximum of $\mu$ degrees of evidence". 

Annotations can be made by $1$ or $n$ values, in order to acquire the correct intensity of the representations regarding how much the annotations expose the knowledge about a proposition $P$.

It is common to use a lattice in the real plane where the annotations take the form of two numeric values ($\mu $,$\lambda $). In this case, the logic is referred to as Paraconsistent Annotated logic with Annotation of Two Values (PAL2v).

 The annotation of two values in the lattice $\tau$ is represented by: $\tau = \{ ( \mu ,\lambda) |\,\mu ,\lambda \in [ 0,1 ] \subset \mathds{R}\}$. The operator “$\thicksim$” can now be defined as: \[\thicksim~[( \mu ,\lambda)] = (\lambda, \mu )\] 

Thus, the association of an annotation $( \mu ,\lambda)$ with a proposition $P$ means that the {\em favourable} evidence degree about $P$ is $\mu $, while its {\em unfavourable} evidence degree is $\lambda $. PAL2v defines the lattice shown in Figure \ref{3}, representing the following:

\begin{itemize}
    \item $P_t$ = $P\left( \mu, \lambda \right )$ = $P\left( 1,0 \right )$: indicating a total favourable evidence and null unfavourable evidence for $P$, assigning a logical connotation of {\em True}.

\item $P_F$ = $P\left( \mu, \lambda \right )$ = $P\left( 0,1 \right )$: indicating a null favourable evidence and total unfavourable evidence for $P$, assigning a logical connotation of {\em False}.

\item $P_\top $ = $P\left( \mu, \lambda \right )$ = $P\left( 1,1 \right )$: indicating a total favourable evidence and a total unfavourable evidence for $P$, thus assigning a connotation of {\em Inconsistency}.

\item $P_\perp $ = $P\left( \mu, \lambda \right )$ = $P\left( 0,0 \right )$: indicating a null favourable evidence and null unfavourable evidence for $P$, assigning a logical connotation of {\em Paracompleteness}.
\end{itemize}

The collection of values for the degrees of favourable evidence ($\mu $) and unfavourable evidence ($\lambda $) aim to solve the problem of contradictory signals, facilitating actions to modify the behaviour of a system in order to decrease the intensity (value) of the contradictory signals.

The PAL2v also considers the definitions for the {\em degree of certainty} ($D_c$) and a {\em degree of contradiction} ($D_{ct}$) as given by the Equations \ref{eq10} and \ref{eq11} below:
\begin{align}
&D_c = \mu - \lambda \label{eq10}\\
&D_{ct} = \left (\mu + \lambda  \right ) -1 \label{eq11}
\end{align}

The degree of certainty ($D_{c}$) and contradiction ($D_{ct}$) are defined on the interval $[-1,1]\in \mathds{R}$.

\begin{figure}[ht!]
\centering
\includegraphics[scale=0.08]{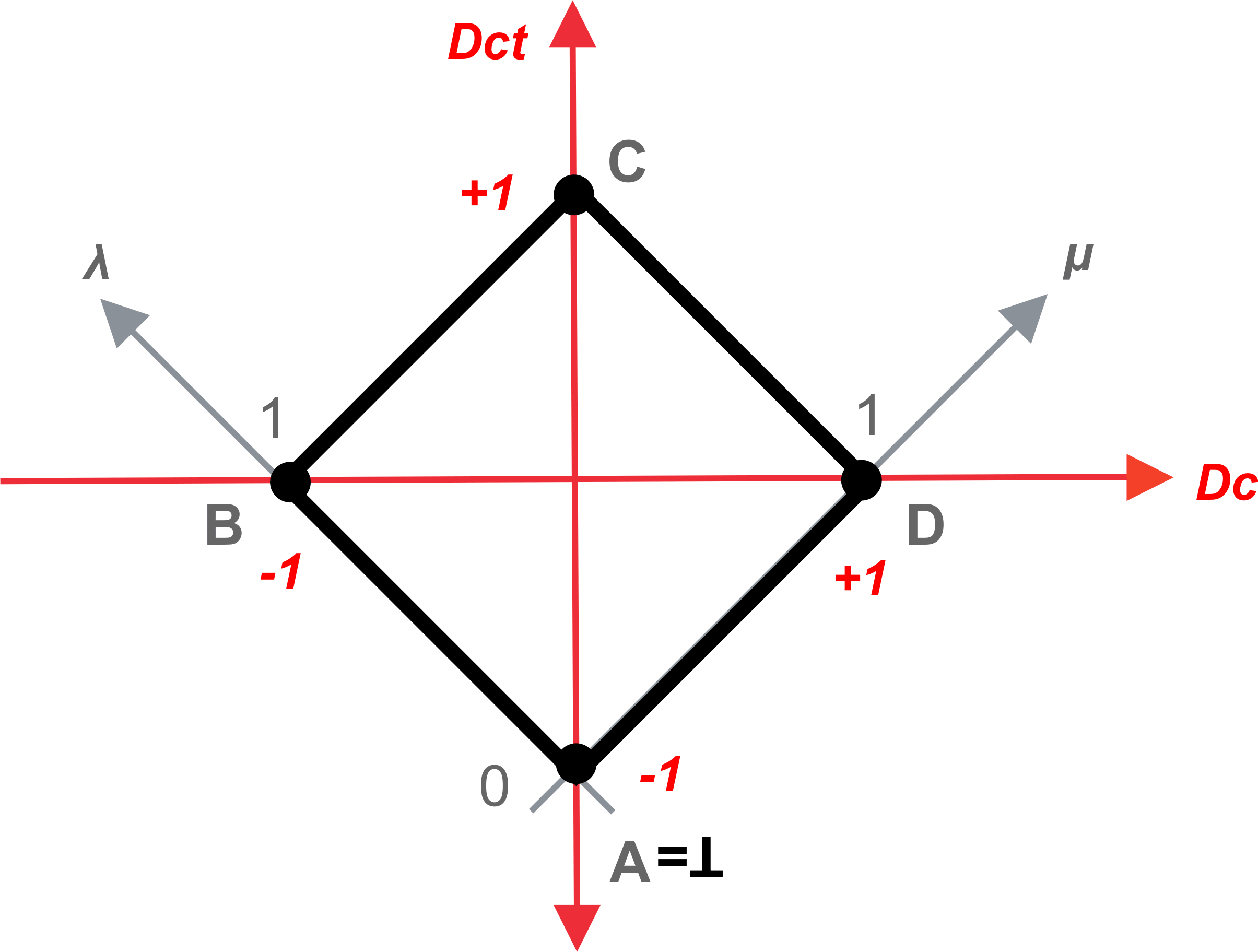}
\caption{Representation of $\mu $, $\lambda $, $D_{c}$ and $D_{ct}$ in the lattice.}\label{4}
\end{figure}

The meeting point between the degree of certainty and the degree of contradiction (as shown in Figure \ref{4}) is considered a paraconsistent logical state $\varepsilon_{\tau}$ = $( D_c,D_{ct}) $. As $\tau$ assumes values closer to the point {\bf D}, paraconsistent logical states  $\varepsilon_{\tau}$ with $D_c$ approximately 1 and with $D_{ct}$ approximately 0 indicate proximity to the truth ($P_t$). In turn, as $\tau$ assumes values closer to the point {\bf B}, the paraconsistent logic states $\varepsilon_{\tau}$ with values $D_c$ approximately -1 and $D_{ct}$ approximately 0 indicate a tendency towards falsehood ($P_F$). Considering that the point of intersection between the degrees of certainty and contradiction axes is the origin of these values, that is, at this point, $D_c$ = 0 and $D_{ct}$ = 0, when propositions present a high contradiction value, for $D_{ct}$ tending to +1 and $D_c$ to 0 (point {\bf A} in Figure \ref{4}), this characterises the logical $\varepsilon_{\tau}$ states that are close to inconsistency ($P_\top $), and the opposite, for $D_{ct}$ approximately -1 and $D_c$ approximately 0 (point {\bf C} in Figure \ref{4}), characterises logical $\varepsilon_{\tau}$ states with proximity to indeterminacy ($P_\perp $).

In the case of a paraconsistent verification system, when a logical state of a proposition remains close to the line segment defined by the {\bf BD} points (Figure \ref{4}), the degree of contradiction in the evidence is small and tends to be null. Similarly, logical whose degree of contradiction is $D_{ct}$ = +1 is an inconsistent state (analogously to the indeterminate state, $D_{ct}$ = -1).

\begin{figure}[ht!]
\centering
\includegraphics[scale=0.07]{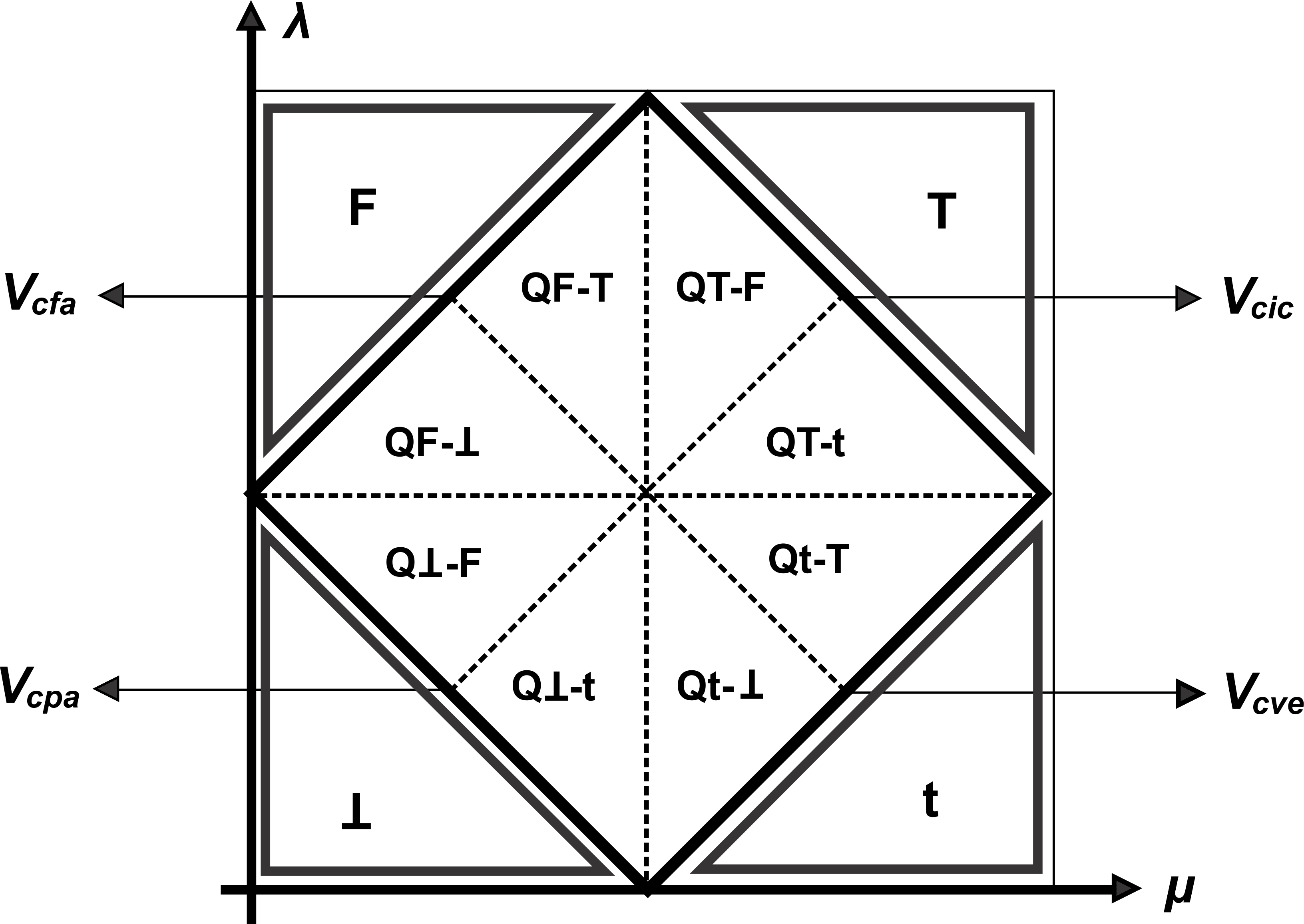}
\caption{Representation of the 12-state lattice subdivision (Para-Analyser)}\label{5}
\end{figure}

With the degrees of certainty and uncertainty, the following 12 output logical states can be defined (as shown in Figure \ref{5}): 
\begin{itemize}
    \item the {\em extreme logical states}: True (t), False (F), Inconsistent($\top $) and Paracomplete($\perp $); 
    \item the {\em Near True tending to Inconsistent} (Qt-$\top $); 
    \item the {\em Near True tending to Paracomplete} (Qt-$\perp $); 
    \item the {\em Near False tending to Inconsistent} (QF-$\top $); 
    \item the {\em Near False tending to Paracomplete} (QF-$\perp $); 
    \item the {\em Near Inconsistent tending to True} (Q$\top $-t); 
    \item the {\em Near Inconsistent tending to False} (Q$\top $-F); 
    \item the {\em Nearly Full Tending to True} (Q$\perp $-t); and,
    \item the {\em Nearly Full Tending to False} (Q$\perp $-F).
\end{itemize}

Figure \ref{5} also shows the boundaries of the regions that define the transition between extreme and non-extreme logical states as: $V_{cic}$ representing the {\em Maximum Uncertainty Control Value}; $V_{cve}$ standing for the {\em Maximum Certainty Control Value}; the $V_{cpa}$, representing the {\em Minimum Uncertainty Control Value}; and, the $V_{cfa}$ representing the {\em Minimum Certainty Control Value}.

Based on the 12-state lattice (Figure \ref{5}), an algorithm called Para-Analyser was created to calculate an output considering the 12 logical states (extremes and non-extremes) as inputs. The pseudocode for this process is shown in Algorithm \ref{alg1} \citep{Abe15ch}.

\begin{algorithm}
\caption{Para-Analyser} \label{alg1}
\begin{algorithmic}[1]

\Procedure{para-analyser}{$mi$ , $la$}       \Comment{Reading of $\mu $ and $\lambda $ through $mi$ and $la$}

    \State $c1 \gets 0.5$;  \Comment{Maximum Certainty Control Value ($Vcve$)}
    \State $c2 \gets -0.5$; \Comment{Minimum Certainty Control Value ($Vcfa$)}
    \State $c3 \gets 0.5$;  \Comment{Maximum Uncertainty Control Value ($Vcic$)}
    \State $c4 \gets -0.5$; \Comment{Minimum Uncertainty control Value ($Vcpa$)}
    
    \State
    
    \State $dc \gets mi - la$;          \Comment{$D_{c}$ calculation}
    \State $dct \gets (mi + la) - 1$;   \Comment{$D_{ct}$ calculation}
    
    \State
    
    \If{($dc \geq c1$)} $s1 \gets t$; \Comment{Section of extreme logical states} \EndIf
    \If{($dc \leq c2$)} $s1 \gets F$; \EndIf
    \If{($dct \geq c3$)} $s1 \gets \top $; \EndIf
    \If{($dct \leq c4$)} $s1 \gets \perp $; \EndIf
    
    \State
    
    \If{($0 \leq dc < c1$) \& ($0 \leq dct < c3 $)} \Comment{Section of non-extreme logical states} \If {($dc \geq dct$)} $s1 \gets Qt-\top$;  \Else \State $s1 \gets Q\top -t$; \EndIf \EndIf
    
    \If{($0 \leq dc < c1$) \& ($c4 < dct \leq 0 $)} \If {($dc \geq dct$)} $s1 \gets Qt-\perp$;  \Else \State $s1 \gets Q\perp -t$; \EndIf \EndIf
    
    \If{($c2 < dc \leq 0$) \& ($c4 < dct \leq 0 $)} \If {($dc \geq dct$)} $s1 \gets QF-\perp$;  \Else \State $s1 \gets Q\perp -F$; \EndIf \EndIf
    \newpage
    \If{($c2 < dc \leq 0$) \& ($0 \leq dct < c3 $)} \If {($dc \geq dct$)} $s1 \gets QF-\top$;  \Else \State $s1 \gets Q\top -F$; \EndIf \EndIf
    
    \State
    
    \State $s2a \gets dc$;
    \State $s2b \gets dct$;
    
    \State
    
    \State \textbf{return}(s1, s2a, s2b); \Comment{Return of current logical state, $D_{c}$ and $D_{ct}$ through $s1$, $s2a$ and $s2b$}
\EndProcedure

\end{algorithmic}
\end{algorithm}

\subsection{Using Fuzzy Sets to obtain paraconsistent logical states}\label{fuzzysets}

This section presents an introduction to the combination of  Fuzzy set theory with Paraconsistent Logics for control systems \citep{Cortes17}.

Degrees of evidence are processed by the Para-Analyser algorithm described above \citep{Abe15ch}, which transforms them into degrees of certainty $D_{c}$ and contradiction $D_{ct}$. These two signals have their values bounded to a normalised interval $[-1,+1]$. This facilitates the application of Fuzzy set strategies for turning each obtained value of $D_{c}$ and $D_{ct}$ into a single output signal that falls in one of the 12 PAL2v states (Figure \ref{5}). This facilitates a finer control of the system, taking into account various degrees of evidence and a rich set of logical states, fine-tuning the decision process of the situation at hand.

\begin{figure}[ht!]
\centering
\includegraphics[scale=0.08]{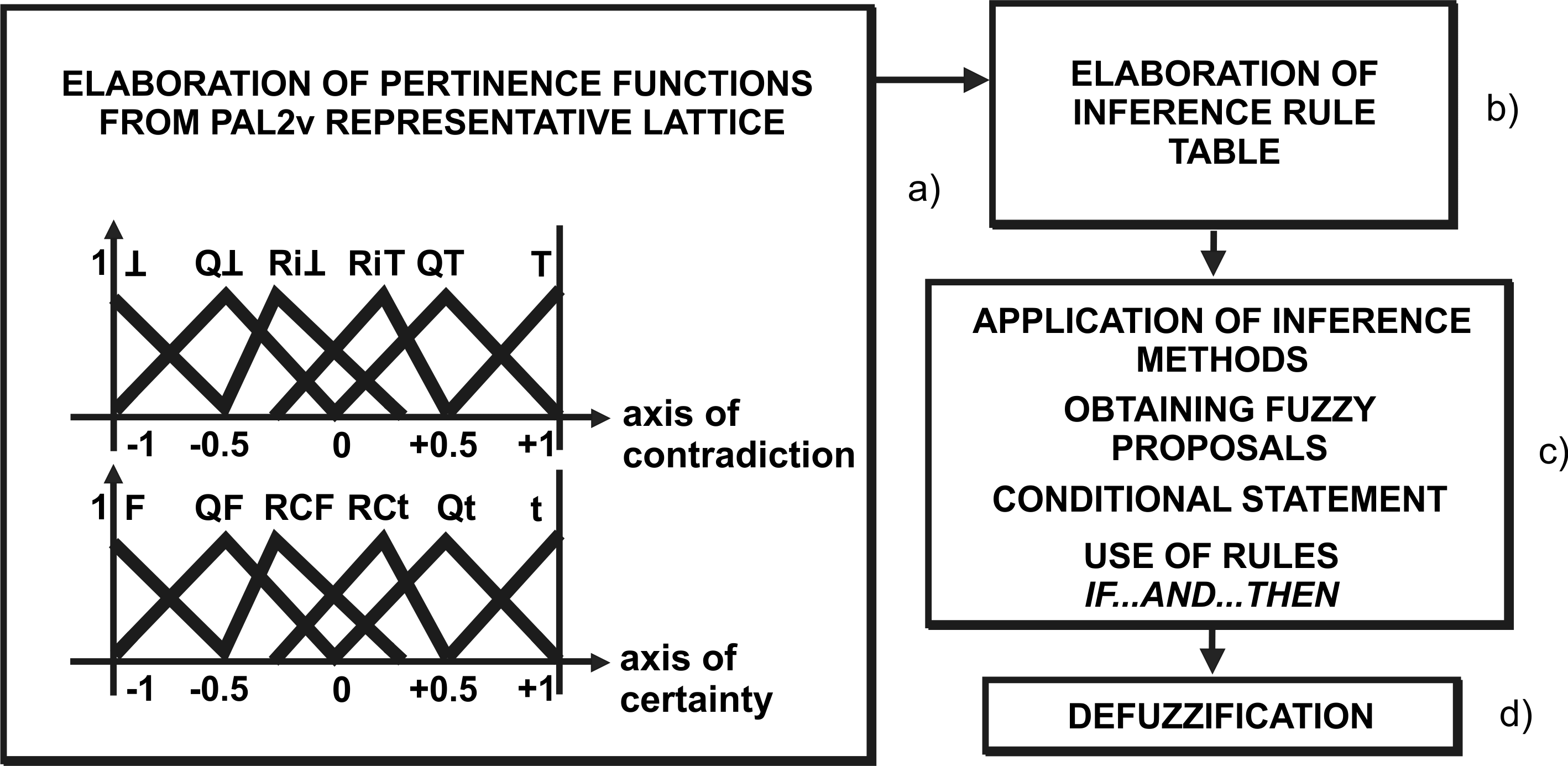}
\caption{Para-fuzzy stages.}\label{6}
\end{figure}

The steps of applying Fuzzy set methods to paraconsistent logical states are summarised in Figure \ref{6}, representing (a) the fuzzification of the degrees of certainty and contradiction; (b) the elaboration of the inference rule table; (c) the application of inference methods; and, (d) defuzzification. 

\begin{figure}[ht!]
\centering
\includegraphics[scale=0.1]{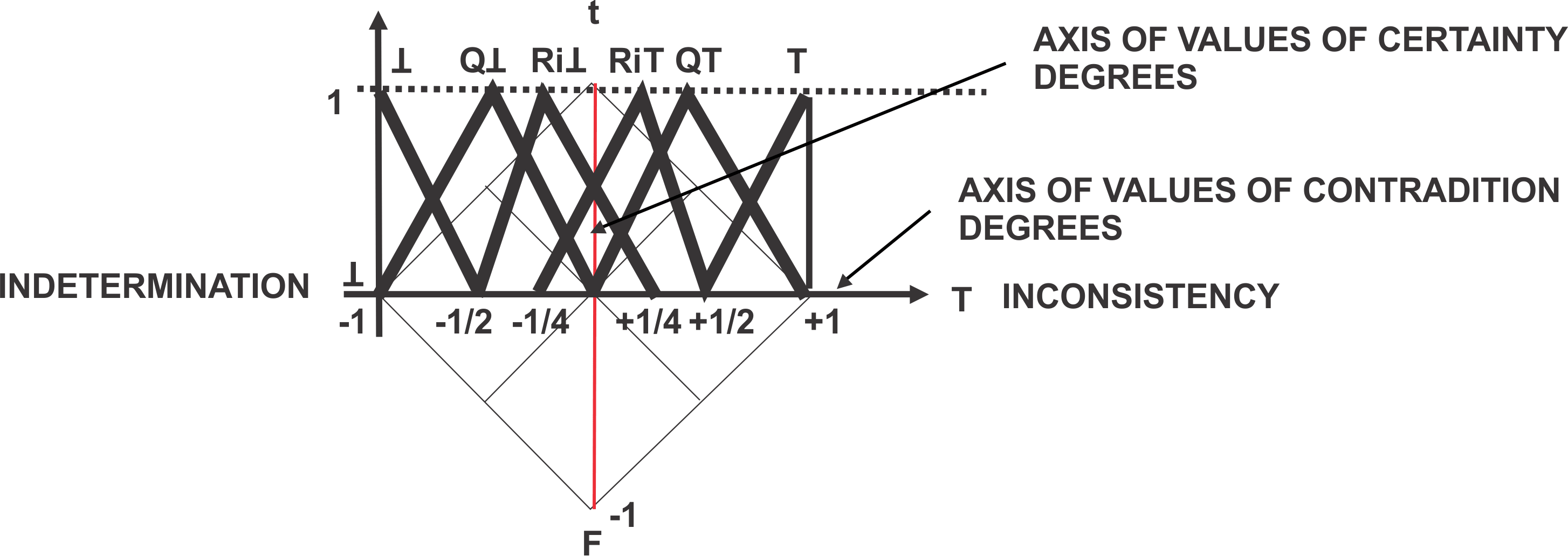}
\caption{Fuzzy membership functions for the contradiction axis.}\label{7}
\end{figure}

\begin{figure}[ht!]
\centering
\includegraphics[scale=0.1]{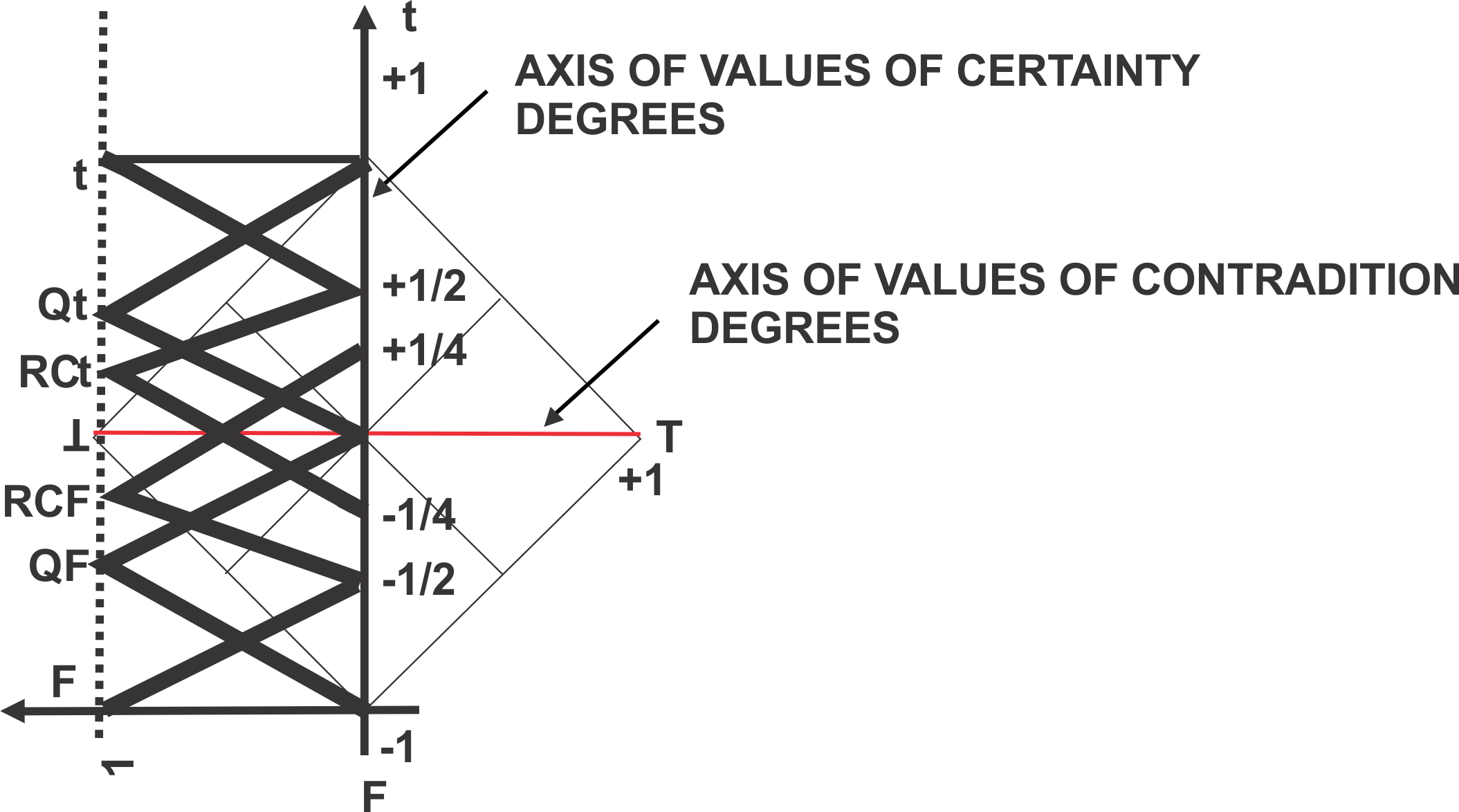}
\caption{Fuzzy membership functions for the certainty axis.}\label{8}
\end{figure}

In step (a) (Figure \ref{6}) the axis representing the degrees of contradiction ($D_{ct}$) and the axis representing the degrees of certainty ($D_{c}$) were fuzzified (as shown in Figures \ref{7} and \ref{8}), using triangular membership functions. These functions were defined in the contradiction axis as: $\perp$ (Indeterminacy), Q$\perp$ (Almost Indeterminate), Ri$\perp$ (Common Region Tending to Undefined) and Ri$\top$ (Common Region Tending to Inconsistent), Q$\top$ (Quasi-Inconsistency) and $\top$ (Inconsistency), and on the certainty axis such as: F (Falsehood), QF (Quasi-Falsehood), RCF (Common Region of Falsehood), RCt (Common Region of Truth) and t (Truth).

\begin{figure}[ht!]
\centering
\includegraphics[scale=0.08]{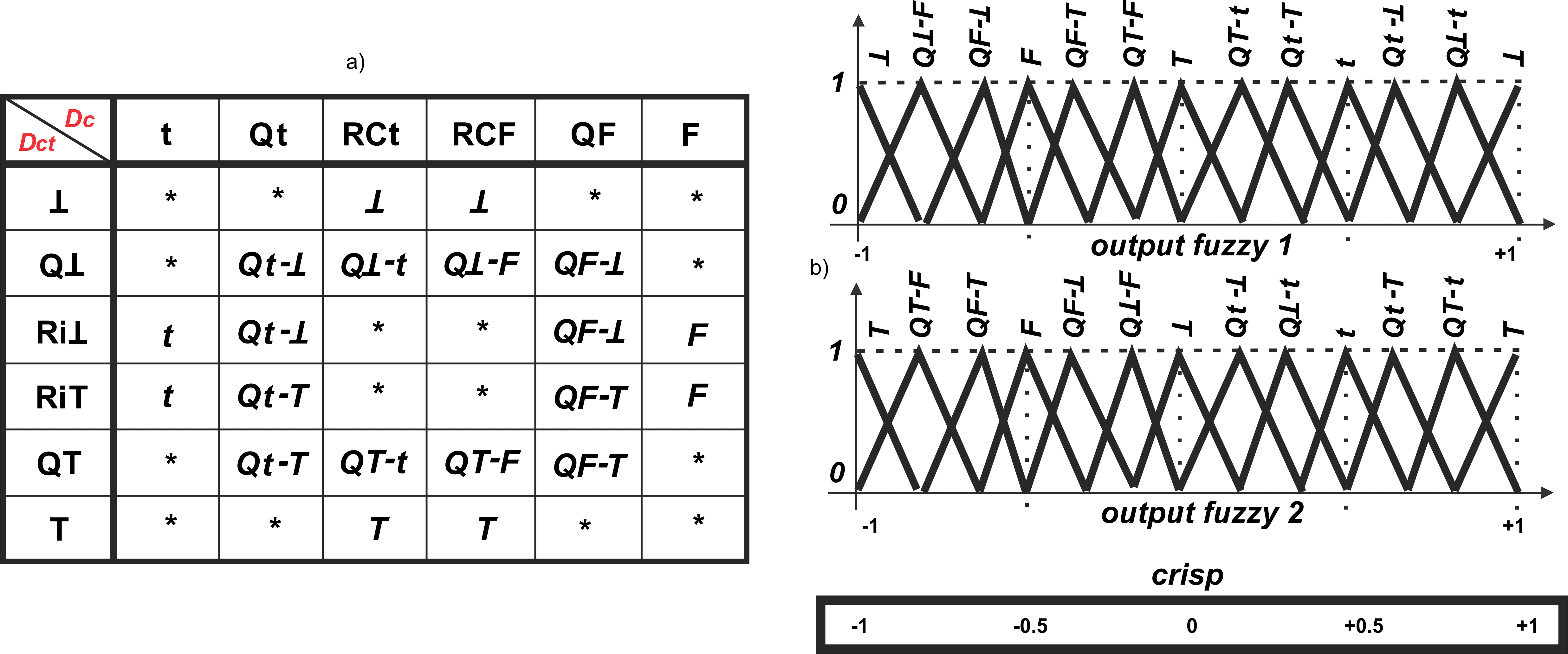}
\caption{Inference rule table, outputs membership functions and crisp.}\label{9}
\end{figure}

Thereafter, by means of correlation between $D_{c}$ and $D_{ct}$, the 12 states of PAL2v (Figure \ref{5}), the pertinence functions (Figures \ref{7} and \ref{8}), and associating the corresponding PAL2v state, the inference table of step (b) was generated. This inference table is shown in Figure \ref{9}a, where the possible combinations of $D_{c}$ and $D_{ct}$, generate PAL2v states, whereas non-possible combinations are represented by ``*" in the table.

The inference rules, from step (c), were performed considering the inference table in Figure \ref{9}a. For step (d), that corresponds to defuzzification, two Fuzzy sets of outputs were used (called  output Fuzzy 1 and  output Fuzzy 2, in Figure \ref{9}b),  to achieve a single output value (crisp) ranging from -1 to +1, that is the result of the combination of paraconsistent logic and Fuzzy logic. 
With this single output value (crisp), it becomes possible to indicate whether a proposition is in one of the 12 regions of the PAL2v lattice (Figure \ref{5}) in a refined way.

The next section presents how this combined Paraconsistent-Fuzzy System was used in the implementation of an expert system for monitoring equipment of a data network in a Brazilian electrical substation.

\section{Paraconsistent-Fuzzy (Para-Fuzzy) Expert System for Data Network Equipment Monitoring of electrical systems}\label{parafuzzy}

This section presents the expert system and subsystems that compose the data network equipment monitoring system developed in this work.

\begin{figure}[ht!]
\centering
\includegraphics[scale=0.06]{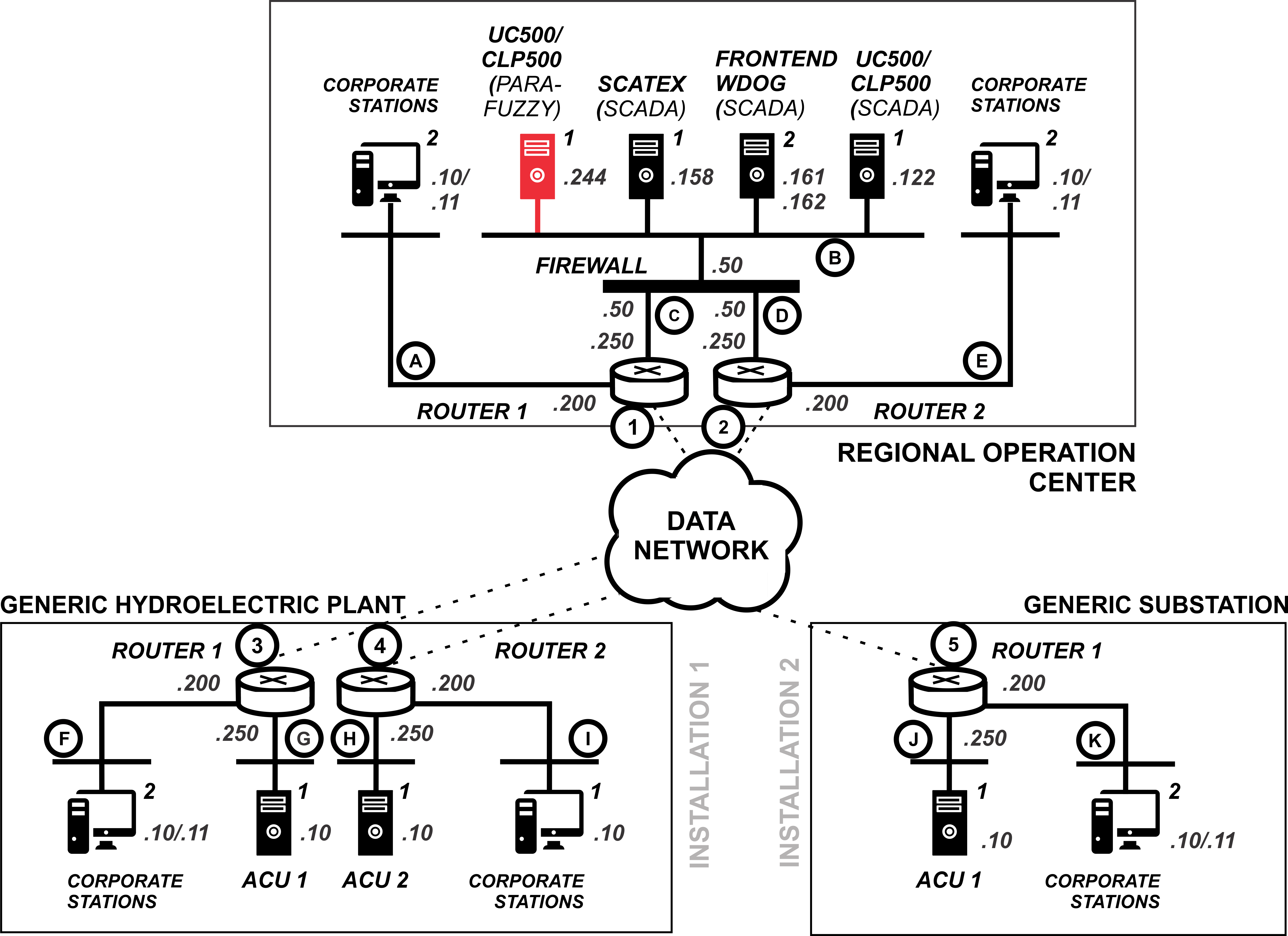}
\caption{Network architecture.}\label{10}
\end{figure}

This research was developed assuming the following hardware and software\footnote{That was available in the power systems laboratory at Santa Cecília University - UNISANTA, where this work was developed.}: HP PROLIANT 360 G8 server, HP WORKSTATION Z240 server and CISCO 2911 router, and Scatex, FE500 5.4.0, CLP500 5.4.0, Wireshark, Matlab R2008a and VMware. With these tools, two test environments (denoted {\em installation 1} and {\em installation 2} below) were constructed by simulating the monitoring of a Regional Operation Centre with the basic Scatex software feature and two remote-controlled facilities. {\em Installation 1} is a Generic Hydroelectric Plant with two Acquisition and Control Units (ACU), three corporate stations and two routers (through which the outgoing network interfaces provide access to the Regional Operation Centre). {\em Installation 2} is a Generic Substation with an ACU, two corporate stations and a router. The Regional Operation Centre was composed of a Scatex Server (with its two aggregated Frontend and Watchdog servers), two UC500 / CLP500 (among them the UC500 / CLP500 Para-Fuzzy Server, where the Para-Fuzzy Expert System was installed), four corporate stations, one firewall, and two routers allowing access to the external facilities to be controlled. This setup is schematised in Figure \ref{10}.

\begin{figure}[ht!]
\centering
\includegraphics[scale=0.055]{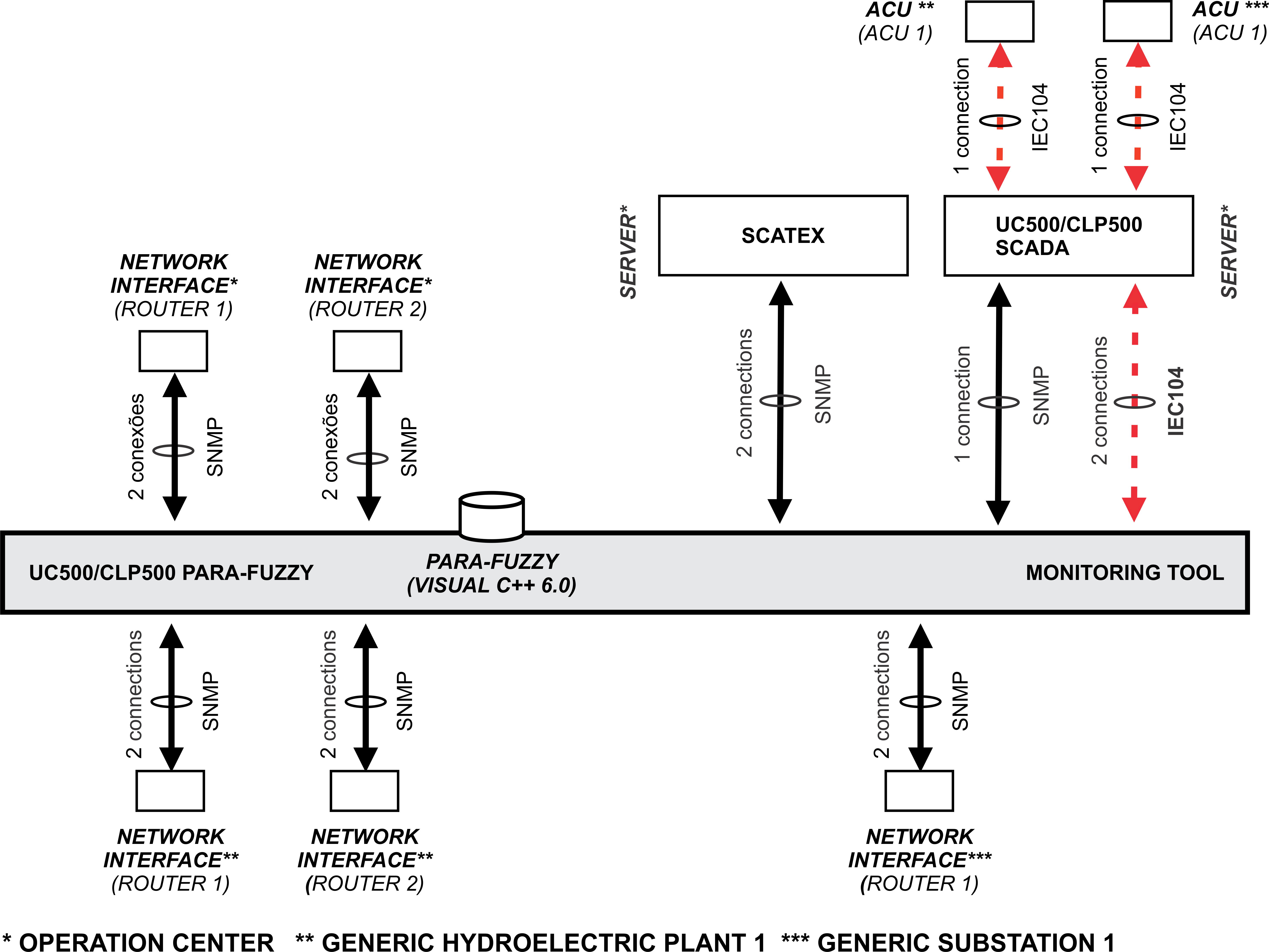}
\caption{Architecture of the expert system for data network equipment monitoring of electrical systems.}\label{11}
\end{figure}

\begin{figure}[ht!]
\centering
\includegraphics[scale=0.06]{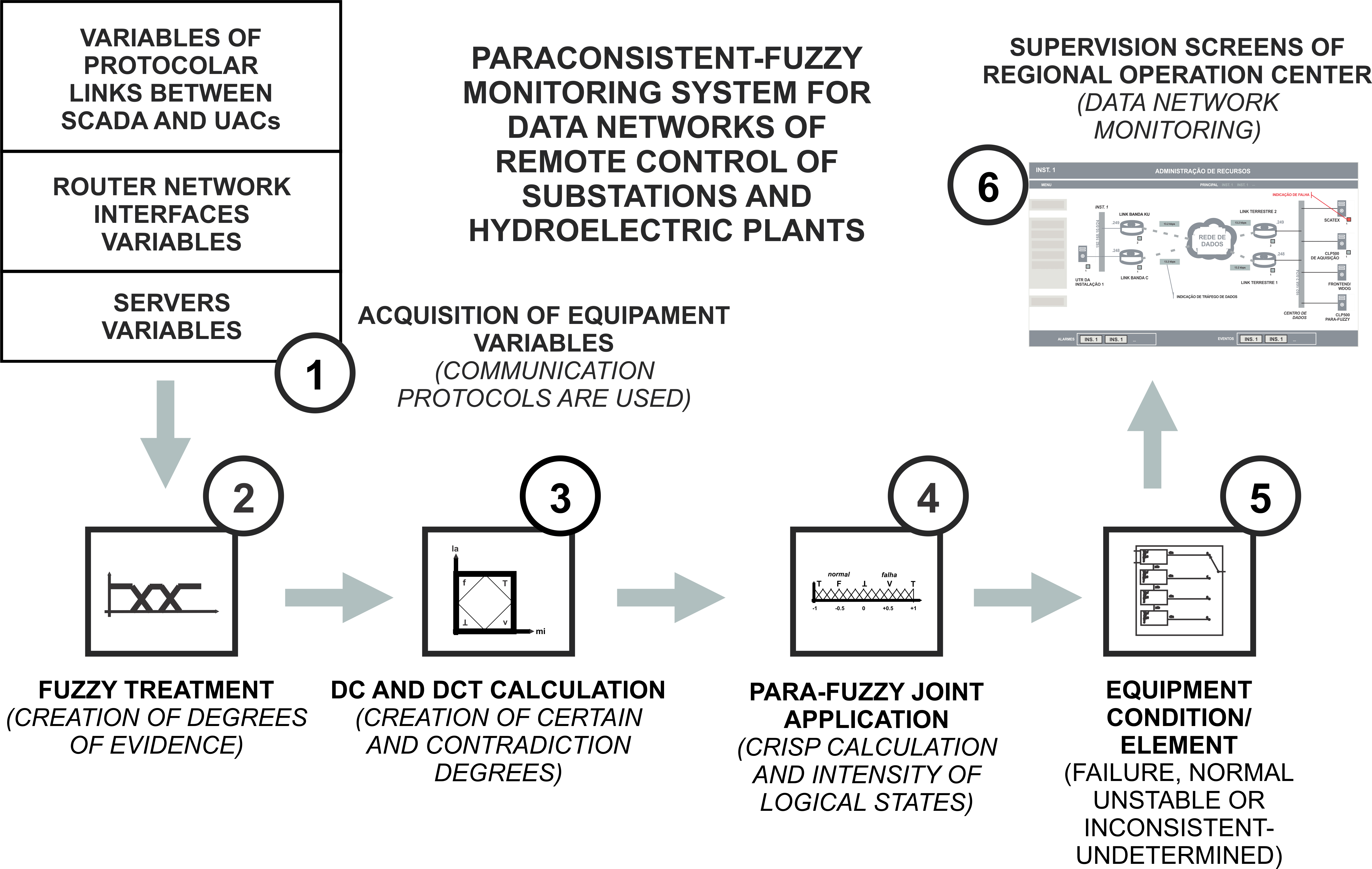}
\caption{General scheme of the Para-Fuzzy Expert System for data network equipment monitoring of electrical systems.}\label{12}
\end{figure}

The subnetwork B of the Regional Operation Centre (Figure \ref{10}) includes the SCADA environment behind the firewall, whereas subnetworks A, C, D and E are connected directly to the routers.  Subnetworks B, C, D, G, H and J communicate in an operative segment (logical segmentation). Subnetworks A, E, F, I, and K communicate and interact with other elements in the data network cloud, and form the enterprise segment. The elements identified by the numbers 1, 2, 3, 4, and 5 are the outbound interfaces of routers that connect locations to the data network cloud, that is, all the information flow {\em in and from} a location passes through one of these interfaces. Data were collected from the network using SNMP and IEC104 channels (shown in Figure \ref{11}) that served as input to the Para-Fuzzy Expert System for data network equipment monitoring of electrical systems developed in this work. A scheme of how the expert system is integrated in the data network is shown in Figure \ref{12}.

The input of the Para-Fuzzy Expert System (step 1 in Figure \ref{12}) are variables representing the equipment status via communication protocols (as represented in Figure \ref{11}). The values obtained were transformed into degrees evidence by means of a Fuzzy procedure (step 2). With these degrees, the parameters $D_{c}$ and $D_{ct}$ (degrees of certainty and contradiction) are calculated (step 3), and they are used by another function (step 4), that combines paraconsistent and Fuzzy procedures, as described in Section \ref{fuzzysets},  that allows the assignment of one of the 12 PAL2v logical states for each equipment analysed. This is then used to infer the operating condition of the equipment, classified into {\em failure, unstable, normal, inconsistent}  or {\em undetermined} (step 5). Information about the entire process is finally sent to the server with Scatex software, so that the automation specialist and the system operators can monitor the network (step 6).

The scheme shown in Figure \ref{12} was used in three subsystems, each of which dedicated to monitoring the: (A) servers; (B) routers' communication; and, (C) ACU connection.

Subsystem (A) receives as input the processing information, as well as the memory status and file allocation, and outputs the operating condition of the analysed server. Subsystem (B) uses the data-rate information and communication-interface errors to assign the operating condition of the analysed router. Similarly, subsystem (C) analyses a particular protocol link from an ACU, using as input the data-rate information and communication interface errors, and infers the operating condition of the analysed ACU.

Steps 2 to 4 (Figure \ref{12}) were summarised into a structure called the {\em analysis node}, this was done for code optimisation and also due to the fact that the paraconsistent-fuzzy combination analyses at most two variables at a time. Thus, in the subsystem (A) two nodes were used and in the subsystems (B) and (C) only one node was used. In the subsystem (A), one node analyses the aggregation of processing and memory, while the other analyses processing and file allocation. In the subsystem (B), the analysis was conducted with the aggregation of the data rate and errors in the router network interface, and in the subsystem (C) the analysis was done with data rate messages and the errors in the protocol connections, given in industrial format via the ACUs.

The expert subsystems run in parallel on a 5-minute loop on the UC500 / CLP500 Para-Fuzzy Server (from Figure \ref{10}). There can be a number of these subsystems for analysing a data network, one for each equipment to be monitored.

The degrees of evidence (favourable and unfavourable) were obtained from raw values using Fuzzy trapezoidal membership functions with three categories representing three levels of failures: {\em not failure}, {\em undefined}, {\em failure}. This allowed the projection of the degrees of evidence in the lattice of the PAL2v, as described in Section \ref{fp-intro},  accommodating the operating condition of the equipment into the states Q$\perp $-F, QF-$\perp $, QF-$\top $,  Q$\top$-F, Q$\top $-t, Qt-$\top $,  Qt-$\perp $ and Q$\perp $-t representing conditions of instability (without sufficient evidence to infer failure), t representing a condition of failure (with evidence of failure), F is a condition of normality (with evidence of non-failure), and $\perp $ and $\top $ are conditions of indeterminacy / inconsistency (evidences with a high level of contradiction).

In the experimental evaluation of the expert system (presented later in this article), the values for the trapezoidal functions and the rules of the Fuzzy procedure of step 2 were defined individually for each expert  subsystem (corresponding to the input and output variables) and for each equipment. These values and rules were defined manually by the electrical-system specialist and are shown in Tables \ref{long}, \ref{long_1} and \ref{long_2} in the appendix. The other Fuzzy procedures in step 4 are those described in Section \ref{fuzzysets}, and also in \citep{inacio99}.

\section{Experimental Setup}\label{exp}

In order to evaluate the proposed expert system, simulations were carried out in MATLAB and in the software CLP500 and Scatex. In the MATLAB simulations, the subsystems described in the previous section were modelled as functions whose inputs were numerical values representing the information obtained by monitoring the various equipment. The outputs of these functions were matrices of values, one matrix for each subsystem, representing the individual operating conditions. For subsystem (A), the matrix represented values for processing, memory, file allocation, $\mu $, $\lambda $, $D_{c}$, $D_{ct}$, $crisp$ and equipment operating condition; for subsystem (B) the output matrix had rate, errors, $\mu $, $\lambda $, $D_{c}$, $D_{ct}$, $crisp$ and equipment operating condition; and for subsystem (C): rate, errors, $\mu $, $\lambda $, $D_{c}$, $D_{ct}$, $crisp$ and equipment operating condition. The value $crisp$ is the control action obtained after the defuzzification procedure (as described in Section \ref{fp-intro}).

In order to conduct the system evaluation using the CLP500 and Scatex software, a local test network was created simulating the following equipment of the architecture shown in Figure \ref{10}: the subnet B servers, the router 1 of the Generic Hydroelectric Plant and the UAC 1 of the Generic Hydroelectric Plant, which used for this test procedure the hardware and software presented at the beginning of Section \ref{parafuzzy}. The simulation of this local network was sufficient to test the three distinct expert  subsystems. In this simulation, functions corresponding to the specialist subsystems were programmed and, on the Scatex Server, information displays were created. Using the connection to the local network, equipment data were collected by protocols and, then, these functions analysed the data and generated the results in tables for the UC500 / CLP500 Para-Fuzzy Server itself and on the displays of the Scatex Server. It is important to highlight that the output tables of the UC500 / CLP500 Para-Fuzzy Server had the same data pattern and they were subjected to the same operating conditions as the MATLAB simulations, allowing a consistent comparison of the results.

\section{Results}\label{res}

The results of the MATLAB simulation are described in Section \ref{comb_results}. Tests were conducted by using as input the whole range of $D_{c}$ and $D_{ct}$ values along the lattice. Two lattices were plotted as results, one representing the crisp values obtained and another highlighting the PAL2v logical state assigned. A PAL2v state assigned to the system was that associated to the highest value of the Fuzzy membership.

Section \ref{esp_ml_results} presents the MATLAB simulations of subsystems (A), (B) and (C). Similar tests with these subsystems were also performed using the CLP500 and the Scatex Server (instead of MATLAB), with the router 1 and the ACU 1 of the Generic Hydroelectric Plant (as described in Section \ref{esp_sf_results}). Each of these devices were subjected to operating conditions of normality, instability and failure.

\subsection{MATLAB simulations of the Para-Fuzzy Expert System}\label{comb_results}

\begin{figure}[ht!]
\centering
\includegraphics[scale=0.055]{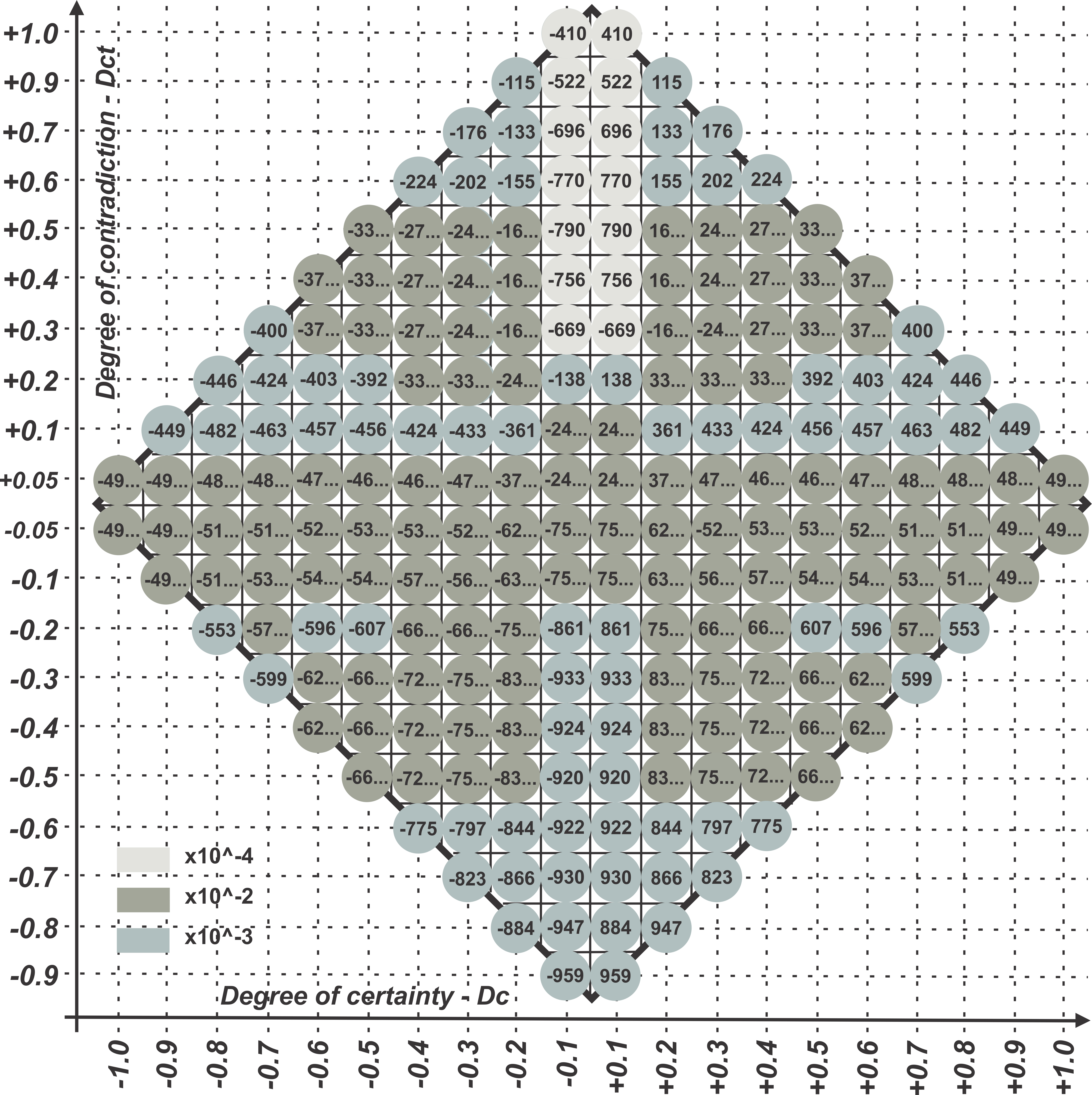}
\caption{Crisp values in the PAL2v lattice.}\label{13}
\end{figure}

\begin{figure}[ht!]
\centering
\includegraphics[scale=0.055]{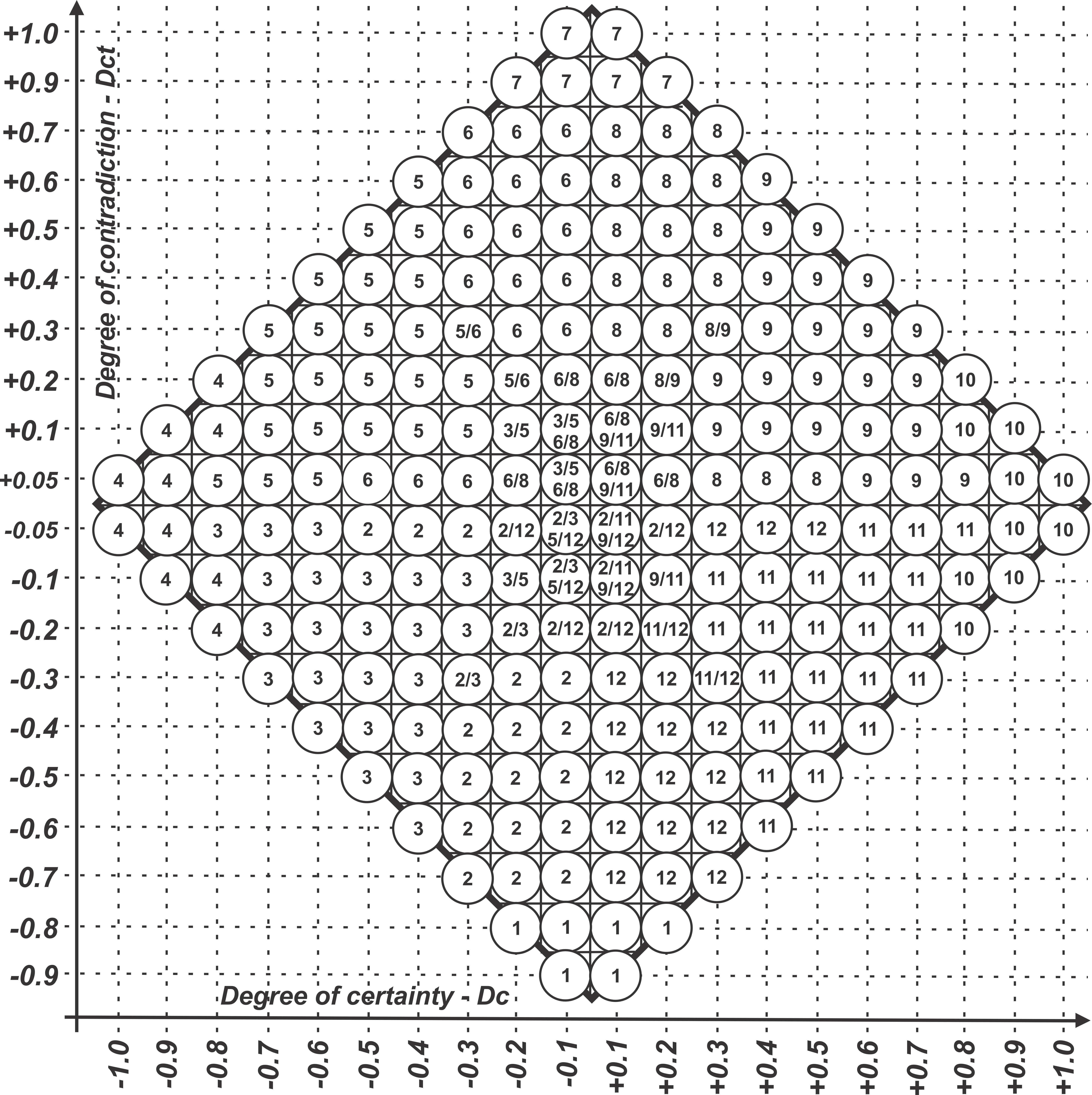}
\caption{Logical states (higher Fuzzy membership value) in the PAL2v lattice.}\label{14}
\end{figure}

Figures \ref{13} and \ref{14} show the results of the paraconsistent and Fuzzy combination presented in Section \ref{fuzzysets}. In Figure \ref{13} , for better representation of the results, the lattice was divided as follows: on the horizontal axis, $D_{c}$ assuming values from $[-1.0$ $to$ $+1.0]$, represented by $[-1.0, -0.9,$ $..., -0.1, -0.05 , +0.05, +0.1, ..., +0.9, +1.0]$, and, in the vertical axis, $D_{ct}$ the values from $[-1.0$ $to$ $+1.0]$, represented by $[-1.0, -0.9, ..., -0.1, +0.1, ..., +0.9, +1.0]$. Still, in Figure \ref{13} the values were represented with different levels of precision, with $10^{-2}$, $10^{-3}$ and $10^{-4}$, and highlighted in different colours. In Figure \ref{14} the 12 PAL2v logical states were represented with numbers 1 to 12, which represent respectively:  $\perp $, Q$\perp $-F, QF-$\perp $, F,  QF-$\top $, Q$\top$-F, $\top $,  Q$\top $-t,  Qt-$\top $, t,  Qt-$\perp $ and Q$\perp $-t.

Figure \ref{13} shows the crisp values\footnote{The unique values obtained as a result of the defuzzification procedure of the paraconsistent-fuzzy combination.} for the different inputs of $D_{c}$ and $D_{ct}$. In these results, it can be seen that when achieving a value of Truth (t), at $D_{c}$ = + 1 and $D_{ct}$ = 0, the crisp values tend to +0.5. In contrast, when tending to False (F), at $D_{c}$ = -1 and $D_{ct}$ = 0, the crisp values tend to -0.5. For high contradictions, high divergence in the measured values for the same situation, Figure \ref{13} shows Inconsistency ($\top $), at $D_{c}$ = 0 and $D_{ct}$ = + 1, and the crisp tending to 0. In the case of Indeterminate ($\perp $), at $D_{c}$ = 0 and $D_{ct}$ = -1, the  crisp values tend to -1 or +1, depending on whether the values come from the right (from the truth) or from the left (from the falsehood).

Figure \ref{14} shows the PAL2v logical states as numbers 1 to 12, for different input values of $D_{c}$ and $D_{ct}$ used in the simulation. The PAL2v logical states are decided according to highest values of the Fuzzy membership function of the combination of inputs, considering a processing interval of 5 minutes. In this figure we can observe that, for a combination of the extreme values for $D_{c}$ (1 or -1), and $D_{ct}$ (+1 or -1) we have true (t) (states with number 10 assigned, in Figure \ref{14}), false (F) (number 4), Inconsistent ($\top $) (number 7) and Indeterminate ($\perp $) (number 1).

\vspace{5mm}

\subsection{Results of the MATLAB simulations for the expert subsystems}\label{esp_ml_results}

Table \ref{long_paper_1} shows the test results of the MATLAB simulation for the expert  subsystem (A), representing the Scatex Server (Figure \ref{11}). Three tests were performed, with the equipment operating in normal condition (test 1), unstable condition (test 2) and failure (test 3). The results for this subsystem were presented for each node (according to the structure presented in Section 3), for better understanding.

In test 1, for node 1 (label (1,1) in Table \ref{long_paper_1}) and  node 2 (label (1,2) in Table \ref{long_paper_1}), a negative certainty value ($D_{c}$) was obtained (-0.8400), whereas the unfavourable evidence level ($\lambda $) was positive (+0.9200) and the crisp value was close to -0.50 (-0.499983866). These values triggered the false logical state (F) of the PAL2v lattice. In this case, the expert system considers as false the statement: ``There is a failure". Thus, confirming that the equipment is in normal working condition.

Test 2 shows the activation of the false logical state (F) for node 2 (label (2,2) in Table \ref{long_paper_1}), for values of $D_c, \lambda$ and $crisp$ similar to those obtained in test 1. However, in this case, node 1 (label (2,1)) shows values for $\mu $ and $\lambda$ that are under 0.5 and, thus, below the threshold to allow the attribution of a categorical truth or false value for the confirming, or not, the statement: ``There is a failure”. Thus, with a low $D_{c}$ value (-0.1218), the triggered states were QF-$\top $ and QF-$\perp $, that correspond to the situations of ``the Near False tending to Inconsistent" and ``the Near False tending to Paracomplete". This situation was, then, considered to be between normal and failure (with crisp values not so close to -0.5 and +0.5), that is, the equipment analysed is in an unstable state.

For test 3, node 1 ((3,1) in Table \ref{long_paper_1}) shows a large positive $D_{c}$ value of +0.8400, and also a large $\mu$ (+0.9200); the crisp value was close to +0.5 (+0.499999996). This situation triggered the truth value (t), according to the PAL2v lattice. Therefore, the statement “There is a failure” is considered true, regardless of the results of node 2 ((3,2) in Table \ref{long_paper_1}).

\vspace{7.5mm}
 \label{my-label-table1}
 {\footnotesize
 \begin{longtable}[h]{ c c c c c c}
 \caption{Tests of the subsystem (A) in the analysis of the Scatex Server (MATLAB simulations)}\label{long_paper_1}\\
 
 \textbf{\shortstack{EQUIP-\\AMENT}}                                    & \textbf{\shortstack{TEST/ \\NODE}}                    & \textbf{\shortstack{ENGINEERING\\ VARIABLES \\READ}}                                  &  \textbf{\shortstack{DEGREES}}                            & \textbf{\shortstack{CRISP AND \\LOGICAL \\ STATE}}       & \textbf{\shortstack{OPERATION \\CONDITION}}       \\
 \shortstack{(description)}                                             & \shortstack{(test number,\\node \\identify)}                 & \shortstack{(P1,P2;M1,M2;\\A1,A2)*}                                            &  \shortstack{($\mu $, $\lambda $;\\$D_{c}$, $D_{ct}$)}    & \shortstack{(crisp;\\logical state)}                     & \shortstack{(equipment \\condition)}              \\
                                                                        &                                                           &                                                                                   &                                                           &                                                           &                                                   \\
 \hline
 \endfirsthead
 
 \multicolumn{6}{c}{Continuation of Table \ref{long_paper_1}}\\
 \hline
 \textbf{\shortstack{EQUIP-\\AMENT}}                                    & \textbf{\shortstack{TEST/ \\NODE}}                    & \textbf{\shortstack{ENGINEERING\\ VARIABLES \\READ}}                                  &  \textbf{\shortstack{DEGREES}}                            & \textbf{\shortstack{CRISP AND \\logical \\ STATE}}       & \textbf{\shortstack{OPERATION \\CONDITION}}       \\
 \shortstack{(description)}                                             & \shortstack{(test numer,\\node \\identify)}                 & \shortstack{(P1,P2;M1,M2;\\A1,A2)*}                                             &  \shortstack{($\mu $, $\lambda $;\\$D_{c}$, $D_{ct}$)}    & \shortstack{(crisp;\\logical state)}                     & \shortstack{(equipment \\condition)}              \\
                                                                        &                                                           &                                                                                   &                                                           &                                                           &                                                   \\
 \hline
 \endhead
 
 \hline
 \endfoot
 
 \hline
 \endlastfoot

 \multirow{4}{*}{\shortstack{SCATEX \\SERVER}}                                                      &                               &                                                                               &                                                   &                                                               &                                           \\
                                                                                                    &   (1,1)                       &   \shortstack{(1\%,0.8\%;20\%\\22\%;60\%,60\%)}                               &   \shortstack{(0.0800,0.9200;\\-0.8400,0.0000)}   &   (-0,499983866;F)                                            &   \multirow{1}{*}{\shortstack{NORMAL}}    \\ \cline{2-5}
                                                                                                    &   (1,2)                       &   \shortstack{(1\%,0.8\%;20\%\\22\%;60\%,60\%)}                               &   \shortstack{(0.0800,0.9200;\\-0.8400,0.0000)}   &   (-0,499983866;F)                                            &                                           \\
                                                                                                    &                               &                                                                               &                                                   &                                                               &                                           \\ \cline{2-6}
                                                                                                    &                               &                                                                               &                                                   &                                                               &                                           \\
 \multirow{8}{*}{\shortstack{SCATEX \\SERVER}}                                                      &   (2,1)                       &   \shortstack{(28\%,28\%;85\%\\85\%;60\%,60\%)}                               &   \shortstack{(0.4392,0.5609;\\-0.1218,0.0000)}   &   \shortstack{(-0,499976939;\\QF-$\perp $ / $\top $)}         &   \multirow{2}{*}{\shortstack{UNSTABLE}}  \\ \cline{2-5}
                                                                                                    &   (2,2)                       &   \shortstack{(28\%,28\%;85\%\\85\%;60\%,60\%)}                               &   \shortstack{(0.0890,0.9110;\\-0.8220,0.0000)}   &   (-0,499979480;F)                                            &                                           \\
                                                                                                    &                               &                                                                               &                                                   &                                                               &                                           \\ \cline{2-6}
                                                                                                    &                               &                                                                               &                                                   &                                                               &                                           \\
                                                                                                    &   (3,1)                       &   \shortstack{(10\%,10\%;95\%\\95.5\%;70\%,70\%)}                             &   \shortstack{(0.9200,0.0800;\\0.8400,0.0000)}    &   (0,499999996;V)                                             &   \multirow{2}{*}{\shortstack{FAIL}}      \\ \cline{2-5}
                                                                                                    &   (3,2)                       &   \shortstack{(28\%,28\%;95\%\\95.5\%;60\%,60\%)}                             &   \shortstack{(0.0800,0.9200;\\-0.8400,0.0000)}   &   (-0,499983866;F)                                            &                                           \\
                                                                                                    &                               &                                                                               &                                                   &                                                               &                                           \\ \cline{2-6}

 \end{longtable}}
 
* P1 - processing reading 1; P2 - processing reading 2; M1 - memory reading 1; M2 - memory reading 2; A1 - reading 1 of file allocation; A2 - reading 2 of file allocation.


Table \ref{long_paper_2} shows the test results of the expert  subsystems (B) and (C) applied, respectively, to the equipment analysis of router 1 and ACU 1 from the Generic Hydroelectric Plant shown in Figure 10. In this case, three tests were also performed for each equipment representing: normal operation, instability and failure. These expert  subsystems were composed of only one node, since there is only one decision, with two variables, for each equipment: rate and error. It is possible to observe in Table \ref{long_paper_2} that, similarly to subsystem (A), in the case of normal operation, $D_{c}$ had a large negative value ($< 0.75$), $\lambda$ had a large positive value ($> 0.75$), and crisp was close to -0.5. This case triggered the false (F) logical state, falsifying the proposition. In the case of failure, the opposite occurred:  $D_{c}> 0.5$, $\mu > 0.5$, and crisp was close to +0.5, which triggered the True (t) logical state for the statement. Under the conditions of instability, $\mu$ and $\lambda$ were between $-0.5$ and $0.5$ that is, in these cases, it was insufficient to assign an absolute truth or false value to the proposition.
 
 \label{my-label-table2}
 {\footnotesize
 \begin{longtable}[h]{ c c c c c c}
 \caption{Tests for subsystems (B) and (C) in the analysis of router 1 and UAC 1 of the Generic Hydroelectric Plant (MATLAB simulations)}\label{long_paper_2}\\
 
 \textbf{\shortstack{EQUIP-\\AMENT}}                                    & \textbf{\shortstack{TEST/ \\NODE}}                    & \textbf{\shortstack{ENGINEERING\\ VARIABLES \\READ}}                                  &  \textbf{\shortstack{DEGREES}}                            & \textbf{\shortstack{CRISP AND \\LOGICAL \\ STATE}}       & \textbf{\shortstack{OPERATION \\CONDITION}}       \\
 \shortstack{(description)}                                             & \shortstack{(test number,\\node \\identify)}                 & \shortstack{(T1,T2;\\E1,E2)}                                            &  \shortstack{($\mu $, $\lambda $;\\$D_{c}$, $D_{ct}$)}    & \shortstack{(crisp;\\logical state)}                     & \shortstack{(equipament \\condition)}             \\
                                                                        &                                                           &                                                                                   &                                                           &                                                           &                                                   \\
 \hline
 \endfirsthead
 
 \multicolumn{6}{c}{Continuation of Table \ref{long_paper_2}}\\
 \hline
 \textbf{\shortstack{EQUIP-\\AMENT}}                                    & \textbf{\shortstack{TEST/ \\NODE}}                    & \textbf{\shortstack{ENGINEERING\\ VARIABLES \\READ}}                                  &  \textbf{\shortstack{DEGREES}}                            & \textbf{\shortstack{CRISP AND \\logical \\ STATE}}       & \textbf{\shortstack{OPERATION \\CONDITION}}       \\
 \shortstack{(description)}                                             & \shortstack{(test numer,\\node \\identify)}                 & \shortstack{(T1,T2;\\E1,E2)}                                            &  \shortstack{($\mu $, $\lambda $;\\$D_{c}$, $D_{ct}$)}    & \shortstack{(crisp;\\logical state)}                     & \shortstack{(equipament \\condition)}             \\
                                                                        &                                                           &                                                                                   &                                                           &                                                           &                                                   \\
 \hline
 \endhead
 
 \hline
 \endfoot
 
 \hline
 \endlastfoot

 \multirow{3}{*}{\shortstack{ROUTER 1\\HYDROELEC.\\PLANT}}                                               &   (1,1)                       &   \shortstack{(900k**,1110k;\\1,0)}                                           &   \shortstack{(0.0800,0.9200;\\-0.8400,0.0000)}   &   (-0,499983866;F)                                            & \shortstack{NORMAL}                       \\ \cline{2-6} \vspace{5pt}
                                                                                                    &   (2,1)                       &   \shortstack{(1300k,1300k;\\9,9)}                                            &   \shortstack{(0.3941,0.6059;\\-0.2118,0.0000)}   &   (-0,499978294;F)                                            & \shortstack{UNSTABLE}                     \\ \cline{2-6} \vspace{5pt}
                                                                                                    &   (3,1)                       &   \shortstack{(0k,0k;\\9,9)}                                                  &   \shortstack{(0.9200,0.0800;\\0.8400,0.0000)}    &   (0,499999996;V)                                             & \shortstack{FAIL}                         \\ \cline{1-6} \vspace{5pt}
 \multirow{3}{*}{\shortstack{ACU 1\\HYDROELEC.\\PLANT}}                                                  &   (1,1)                       &   \shortstack{(1000msg***,990msg;\\0,0)}                                      &   \shortstack{(0.0800,0.9200;\\-0.8400,0.0000)}   &   (-0,499983866;F)                                            & \shortstack{NORMAL}                       \\ \cline{2-6} \vspace{5pt}
                                                                                                    &   (2,1)                       &   \shortstack{(1000msg,990msg;\\0,0)}                                         &   \shortstack{(0.5331,0.4669;\\0.0662,0.0000)}    &   (-0,499999994;F)                                            & \shortstack{UNSTABLE}                     \\ \cline{2-6} \vspace{5pt}
                                                                                                    &   (3,1)                       &   \shortstack{(31msg,30msg;\\0,0)}                                            &   \shortstack{(0.9200,0.0800;\\0.8400,0.0000)}    &   (0,499999996;V)                                             & \shortstack{FAIL}                         \\

 \end{longtable}}
 
 ** number of kbits / 300s; *** number of messages / 300s; T1 - data rate reading 1; T2 - data rate reading 2; E1 - error rate reading 1; E2 - error rate reading 2.
 
 \vspace{2mm}

It is worth noting that all the tests presented in Tables \ref{long_paper_1} and \ref{long_paper_2} were made with readings of variables (processing, memory, etc) that did not cause high values for $D_{ct}$. If this had happened, we would have the Inconsistent condition (for $D_{ct}$ $> 0.5$) and Undetermined ( for $D_{ct}$ $<-0.5$).

\subsection{Results of of the Scatex and CLP500 implementation of the expert subsystems}\label{esp_sf_results}

Tables \ref{long_paper_3} and \ref{long_paper_4} present analogous results to the  test sequence presented in the previous section, but obtained using the Scatex and CLP500 software. These systems are responsible for monitoring and analysing the equipment. They are allocated, respectively, to the servers .244 and .158, of the architecture shown in Figure \ref{10} in a real electrical installation. The results obtained in these simulations were analogous to those presented in the previous section and the system's behaviours received the same classification for similar cases, as shown in Tables \ref{long_paper_3}, \ref{long_paper_4}.

 \label{my-label-table3}
 {\footnotesize
 \begin{longtable}[h]{ c c c c c c}
 \caption{Tests of the subsystem (A) in the analysis of the Scatex Server (in Scatex and CLP500 software)}\label{long_paper_3}\\
 
 \textbf{\shortstack{EQUIP-\\AMENT}}                                    & \textbf{\shortstack{TEST/ \\NODE}}                    & \textbf{\shortstack{ENGINEERING\\ VARIABLES \\READ}}                                  &  \textbf{\shortstack{DEGREES}}                            & \textbf{\shortstack{CRISP AND \\LOGICAL \\ STATE}}       & \textbf{\shortstack{OPERATION \\CONDITION}}       \\
 \shortstack{(description)}                                             & \shortstack{(test numer,\\node \\identify)}                 & \shortstack{(P1,P2;M1,M2;\\A1,A2)}                                          &  \shortstack{($\mu $, $\lambda $;\\$D_{c}$, $D_{ct}$)}    & \shortstack{(crisp;\\logical state)}                     & \shortstack{(equipament \\condition)}             \\
                                                                        &                                                           &                                                                                   &                                                           &                                                           &                                                   \\
 \hline
 \endfirsthead
 
 \multicolumn{6}{c}{Continuation of Table \ref{long_paper_3}}\\
 \hline
 \textbf{\shortstack{EQUIP-\\AMENT}}                                    & \textbf{\shortstack{TEST/ \\NODE}}                    & \textbf{\shortstack{ENGINEERING\\ VARIABLES \\READ}}                                  &  \textbf{\shortstack{DEGREES}}                            & \textbf{\shortstack{CRISP AND \\LOGICAL \\ STATE}}       & \textbf{\shortstack{OPERATION \\CONDITION}}       \\
 \shortstack{(description)}                                             & \shortstack{(test number,\\node \\identify)}                 & \shortstack{(P1,P2;M1,M2;\\A1,A2)*}                                             &  \shortstack{($\mu $, $\lambda $;\\$D_{c}$, $D_{ct}$)}    & \shortstack{(crisp;\\logical state)}                     & \shortstack{(equipment \\condition)}             \\
                                                                        &                                                           &                                                                                   &                                                           &                                                           &                                                   \\
 \hline
 \endhead
 
 \hline
 \endfoot
 
 \hline
 \endlastfoot

 \multirow{4}{*}{\shortstack{SCATEX \\SERVER}}                                                      &                               &                                                                               &                                                       &                                                               &                                           \\
                                                                                                    &   (1,1)                       &   \shortstack{(1\%,0.8\%;20\%\\22\%;60\%,60\%)}                               &   \shortstack{(0.0833,0.9166;\\-0.8333,-2.23E-08)}    &   (-0,499982;F)                                               &   \multirow{2}{*}{\shortstack{NORMAL}}    \\ \cline{2-5}
                                                                                                    &   (1,2)                       &   \shortstack{(1\%,0.8\%;20\%\\22\%;60\%,60\%)}                               &   \shortstack{(0.0833,0.9166;\\-0.8333,-2.23E-08)}    &   (-0,499982;F)                                               &                                           \\
                                                                                                    &                               &                                                                               &                                                       &                                                               &                                           \\ \cline{2-6}
 \multirow{8}{*}{\shortstack{SCATEX \\SERVER}}                                                      &                               &                                                                               &                                                       &                                                               &                                           \\
                                                                                                    &   (2,1)                       &   \shortstack{(28\%,28\%;85\%\\85\%;60\%,60\%)}                               &   \shortstack{(0.4391,0.5609;\\-0.1218,0.0000)}       &   \shortstack{(-0,499979939;\\QF-$\perp $ / $\top $)}         &   \multirow{2}{*}{\shortstack{UNSTABLE}}  \\ \cline{2-5}
                                                                                                    &   (2,2)                       &   \shortstack{(28\%,28\%;85\%\\85\%;60\%,60\%)}                               &   \shortstack{(0.0890,0.9110;\\-0.8220,0.0000)}       &   (-0,499979480;F)                                            &                                           \\
                                                                                                    &                               &                                                                               &                                                       &                                                               &                                           \\ \cline{2-6}
                                                                                                    &                               &                                                                               &                                                       &                                                               &                                           \\
                                                                                                    &   (3,1)                       &   \shortstack{(10\%,10\%;95\%\\95.5\%;70\%,70\%)}                             &   \shortstack{(0.9200,0.0800;\\0.8400,0.0000)}        &   (0,499999996;V)                                             &   \multirow{2}{*}{\shortstack{FAIL}}      \\ \cline{2-5}
                                                                                                    &   (3,2)                       &   \shortstack{(28\%,28\%;95\%\\95.5\%;60\%,60\%)}                             &   \shortstack{(0.0800,0.9200;\\-0.8400,0.0000)}       &   (-0,499983866;F)                                            &                                           \\
                                                                                                    &                               &                                                                               &                                                       &                                                               &                                           \\ \cline{2-6}

 \end{longtable}}
 

 \label{my-label-table4}
 {\footnotesize
 \begin{longtable}[h]{ c c c c c c}
 \caption{Tests of subsystems (B) and (C) in the analysis of router 1 and UAC 1 of the Generic  Hydroelectric Plant (in Scatex and CLP500 software)}\label{long_paper_4}\\
 
 \textbf{\shortstack{EQUIP-\\AMENT}}                                    & \textbf{\shortstack{TEST/ \\NODE}}                    & \textbf{\shortstack{ENGINEERING\\ VARIABLES \\READ}}                                  &  \textbf{\shortstack{DEGREES}}                            & \textbf{\shortstack{CRISP AND \\LOGICAL \\ STATE}}       & \textbf{\shortstack{OPERATION \\CONDITION}}       \\
 \shortstack{(description)}                                             & \shortstack{(test number,\\node \\identify)}                 & \shortstack{(T1,T2;\\E1,E2)}                                                    &  \shortstack{($\mu $, $\lambda $;\\$D_{c}$, $D_{ct}$)}    & \shortstack{(crisp;\\logical state)}                     & \shortstack{(equipment \\condition)}             \\
                                                                        &                                                           &                                                                                   &                                                           &                                                           &                                                   \\
 \hline
 \endfirsthead
 
 \multicolumn{6}{c}{Continuation of Table \ref{long_paper_4}}\\
 \hline
 \textbf{\shortstack{EQUIP-\\AMENT}}                                    & \textbf{\shortstack{TEST/ \\NODE}}                    & \textbf{\shortstack{ENGINEERING\\ VARIABLES \\READ}}                                  &  \textbf{\shortstack{DEGREES}}                            & \textbf{\shortstack{CRISP AND \\LOGICAL \\ STATE}}       & \textbf{\shortstack{OPERATION \\CONDITION}}       \\
 \shortstack{(description)}                                             & \shortstack{(test number,\\node \\identify)}                 & \shortstack{(T1,T2;\\E1,E2)*}                                                   &  \shortstack{($\mu $, $\lambda $;\\$D_{c}$, $D_{ct}$)}    & \shortstack{(crisp;\\logical state)}                     & \shortstack{(equipment \\condition)}             \\
                                                                        &                                                           &                                                                                   &                                                           &                                                           &                                                   \\
 \hline
 \endhead
 
 \hline
 \endfoot
 
 \hline
 \endlastfoot

 \multirow{3}{*}{\shortstack{ROUTER 1\\HYDROELEC.\\PLANT}}                                               &   (1,1)                       &   \shortstack{(900k,1110k;\\1,0)}                                             &   \shortstack{(0.0833,0.9166;\\-0.833333,2.23E-08)}       &   (-0,499982;F)                                               & \shortstack{NORMAL}                       \\ \cline{2-6} \vspace{5pt}
                                                                                                    &   (2,1)                       &   \shortstack{(1300k,1300k;\\9,9)}                                            &   \shortstack{(0.3625,0.5988;\\-0.236325,-0.03867)}       &   (-0,556207;F)                                               & \shortstack{UNSTABLE}                     \\ \cline{2-6} \vspace{5pt}
                                                                                                    &   (3,1)                       &   \shortstack{(0k,0k;\\9,9)}                                                  &   \shortstack{(0.9200,0.0800;\\0.8400,0.0000)}            &   (0,499999996;V)                                             & \shortstack{FAIL}                         \\ \cline{1-6} \vspace{5pt}
 \multirow{3}{*}{\shortstack{ACU 1\\HYDROELEC.\\PLANT}}                                                  &   (1,1)                       &   \shortstack{(1000msg**,990msg;\\0,0)}                                       &   \shortstack{(0.0833,0.9166;\\-0.833333,2.23E-08)}       &   (-0,499982;F)                                               & \shortstack{NORMAL}                       \\ \cline{2-6} \vspace{5pt}
                                                                                                    &   (2,1)                       &   \shortstack{(1000msg,990msg;\\0,0)}                                         &   \shortstack{(0.5328,0.5321;\\0.0007332,0.06495)}        &   (-0,494548;F)                                               & \shortstack{UNSTABLE}                     \\ \cline{2-6} \vspace{5pt}
                                                                                                    &   (3,1)                       &   \shortstack{(31msg,30msg;\\0,0)}                                            &   \shortstack{(0.9166,0.0833;\\0.833333,2.23E-08)}        &   (0,500;V)                                                   & \shortstack{FAIL}                         \\

 \end{longtable}}
 
\section{Discussion}\label{disc}

Figures \ref{13} and \ref{14} show that, as $D_{c}$ $> 0.75$, there is a gradual increase of trust in assigning the logical state true (t) to the system, with the crisp value tending to +0.5. In contrast,  as $D_{c}$ $< -0.75$ the false logical state (F) is inferred, with the crisp value tending to -0.5 . Furthermore, for $D_{ct}$ $> 0.75$ or $D_{ct}$ $< -0.75$ there is the gradual acceptance of $\top $ or $\perp $, and the approximation of crisp to 0, or -1 (on the left-hand side of the diagram) and +1 (on the right-hand side of the diagram).

For $D_{c} > 0$ and $D_{c} < 0.75$, and also for $D_{ct} < 0$ and $D_{ct} > -0.75$, if $D_{c} > D_{ct}$, the logical state Qt-$\perp $ is assigned. As for the opposite, for $D_{c}>D_{ct}$, $Q\perp$-t is assigned to the system, with the crisp assuming values in the interval [0.50, 0.80]. For $D_{c} > 0$ and $D_{c} < 0.75$, and $D_{ct} > 0$ and $D_{ct} < 0.75$, with $D_{c} > D_{ct}$, the system is classified as Qt-$\top $;  Q$\top $-t (with the crisp assuming values in the interval [0.20,0.50]). This occurs also when $D_{c} > 0$ and $D_{c} < 0.75$ and $D_{ct} < 0.75$, but with $D_{c} < D_{ct}$. For $D_{c} < 0$ and $D_{c} > -0.75$ (on the left side of the lattice), the opposite occurs, with the assignment of the respective logical states and crisp.

These results (in numbers and logical states) prove the correct implementation of the paraconsistent-fuzzy combination, since the values obtained in the simulations agree with whose of the general theory presented in Section \ref{fuzzysets}.

With the association of Fuzzy to the determination of logical states t, F, $\perp $ and $\top $, the affirmation or not of a proposition occurs in a two-tier way: observing the pertinence functions of the competing logical states, and defining the output based on the highest value of pertinence. This contrasts with the previous work presented in \citep{inacio99,Abe15ch} that use the Para-Analyser algorithm without the application of Fuzzy sets.

In relation to the tests of the expert  subsystems in the classification of the operating condition of the different equipment tested, it was possible to see that: although there were differences in the values obtained in the MATLAB with respect to the CLP500 e Scatex implementation, the decision obtained by the expert system was consistent in both cases, for the same situations.

In this work, just a limited number of variables were used, but that is not a physical limitation, as more nodes could be implemented in the expert system. The set of variables used was selected based on the minimum  number of nodes necessary for the remote supervision of a large electrical installation.

Finally, it is important to highlight that the underlying formalism assumed from the start the existence of contradictions between the readings of the engineering variables. This work has shown that, even in stated with contradictory signs, the classification was made in a consistent way. Moreover, no significant distinction was observed between the MATLAB simulation and the equivalent implementation in the softwares CLP500 and Scatex.

 \section{Related Works}\label{related}

\cite{pimenta15} describes a way of analysing a communication network using PAL2v from device-network request logs. These logs were composed of four factors, which are: average response time, standard deviation, average packet size and total transactions, which were collected for five days over three periods (morning, afternoon and evening).  These factors were modelled in paraconsistent logic, individually, through degrees of evidence favourable and unfavourable. As the interpretation of these factors in isolation did not lead to a satisfactory conclusion, a global analysis, using paraconsistent logic, was conducted considering the favourable and unfavourable evidence of each factor, multiplied by respective weights to finally diagnose and analyse the behaviour of the network.

\cite{pimenta16} showed a methodology with PAL2v for analysing and detecting problems in a computer network of a public university with 200 computers. In this methodology, from a proxy, parameters of the network were collected and the response time, data volume, bandwidth, number of requests and re-transmissions were calculated. These parameters were then transformed into favourable and unfavourable evidence related to anomaly propositions, which are then dealt appropriately by assigning states in a lattice, interpreting them as the behaviours of the analysed network.

\cite{pimenta18} presented a network anomaly detection methodology, which used as a source of information for analysis, logs from a proxy for a custom router, called a Squid proxy. In the information acquired by the logs, PAL2v concepts were applied to each equipment analysed, with favourable and unfavourable evidence determined according to its parameters. With the aid of a traffic analyser, it was possible to determine the behaviour of the hosts within a specific interval, daily. Considering various logical states, it was possible to modulate a general analysis corresponding to the detection of network anomalies, considering the behaviour of hosts outside the previously stipulated profile.

Similar to that carried out in \citep{pimenta15}, \citep{pimenta16} and \citep{pimenta18}, in the present paper, the measured parameters of network equipment (servers, routers and ACUs) are associated with favourable and unfavourable evidence, and the PAL2v is used for characterisation of failures in data networks. However, in addition to PAL2v, Fuzzy logic is used which provides the identification network states in a more refined way.

There is also work on the applications of PAL2v  based on the traffic-profile analysis of Digital Signatures of Network Segment using Flow Analysis (DSNSF). \cite{pena14} presented a network-anomaly detection approach employing DSNSF, generated with an Autoregressive Integrated Moving Average  (ARIMA) model. In addition, a functional algorithm based on PAL2v was proposed, with the objective of avoiding high false alarm rates, due to traffic variations beyond the expected profile, but identifying the behaviour of the traffic patterns that really harm the network services.

In \cite{pena17}, a methodology for detecting anomalies in computer networks by means of DSNSF was proposed with the help of PAL2v. In this study, DSNSFs are organised under two distinct models, ARIMA and Ant Colony Optimisation for Digital Signature (ACODS). Through the analysis of traffic records, each structure of a DSNSF was used as a standard for the observed traffic resources. A Correlational Paraconsistent Machine (CPM) was then created, based on LPA2v, with the objective of assimilating the DSNSFs of both models and the traffic disturbances caused by network anomalies, which by comparing them with the use of paraconsistent concepts determines the anomaly in a more adequate way, with the consideration of uncertainties in the information. The experimental results of a real assessment of traffic monitoring suggested that CPM responses indicated the possibility of improvement in anomaly detection rates.

In contrast to the work presented in \citep{pena14} and \citep{pena17}, in which the behaviour of the network traffic modelled over time is considered as the default pattern that defines normal functioning, in the present paper the network behaviour is defined within fixed ranges for all the measured parameters of each equipment in the network, taking into account the conditions of normality, instability and failure of each one, regardless of specific antecedent temporal associations and computational resources of historical data. These value ranges are defined by a specialist. 

Related work on Fuzzy logic is described in \citep{olajubu13}, where a scheme for the predictive maintenance in communication networks is proposed using Fuzzy logic with real-time data monitoring. The model combines efficiency with reasonable cost, and works well in modern networks with heterogeneous elements, including: memory, processing cores and hard disk. The collected information is sent to a Data Analysis (DA) system, which analyses the collected information and advises the network administrator on the condition of the network element. The DA is modelled using the Matlab Fuzzy-logic toolbox, and the data used for the Fuzzy-logic model  was collected from data sheets from different component / equipment suppliers. To verify the feasibility of the proposed model, an {\em ad-hoc} network of ten computers was created and evaluated.

In \cite{kotenco17}, a new approach was proposed for the monitoring of Network Elements (NE), in a Multi-Service Network (MSN), based on the application of diffuse logical inference, in which the need to use Fuzzy methods was due to three factors: (1) uncertainty on the causes that can result in node failures in the communication channels; (2) incomplete information about the status of NEs and MSN as a whole, which are subject to processing; (3) delay in the transmission of NE state data to the processing nodes. 

A combination of Fuzzy logic with neural networks in presented in \citep{rudrusamy13}, where a Fuzzy-based diagnostic system was proposed to recognise and identify network operation anomalies, using neural network as a tuner. The focus was on building a Fuzzy system for handling decoded data packets to identify anomalies. Takagi Sugeno's Fuzzy model was used in the implementation of the system, which allowed the detection of network operation anomalies by their intensity, and the ability to choose the appropriate type of alerts. This combination of the diffuse model with the neural network minimised the number of interruptions of the network operators, relieving them from analysing possible false problems.

It is observed that Fuzzy logic has been used in the treatment and analysis of the operation and maintenance of equipment and systems in electrical network systems for a long time, as can be seen in \citep{tomsovic99} and \citep{haiwen06}. In these cases, the expert knowledge is used to create Fuzzy rules for efficient inferences, facilitating the equipment control.

Analogous to the work presented in \citep{olajubu13} and \citep{kotenco17}, the present paper used Fuzzy logic to monitor and identify network equipment failures by analysing the intrinsic resources of network elements (processing, memory and allocation of disk), basically using rules defined by specialists. Additionally, this paper uses the logic PAL2v to represent the contradiction between network parameter measurements and to conduct a real-time diagnosis of failures.

Similar to the work presented in \citep{tomsovic99} and \citep{haiwen06}, this paper used the knowledge from specialists to model critical networks of SCADA systems in the context of the remote control of electrical systems, creating Fuzzy association rules and parameterisation of membership functions,  modelling  equipment failures of these types of data networks.

 In a more general way, the work reported in \citep{pimenta15,pimenta16,pimenta18}, \citep{pena14, pena17} and \citep{rudrusamy13} model the network traffic, and its aggregated subparameters (such as errors, latency, etc.), in a way to indicate anomalies in data networks; PAL2v is used to represent contradiction, as an additional resource for the correct indication of failure. In \citep{olajubu13,kotenco17}, it is observed that the specific resources of network equipment, such as: memory, processing, hard disk, temperature, etc., were modelled according to specialist knowledge as Fuzzy expert systems but without considering contradictions.

\section{Conclusion}\label{conclu}
\label{conclusion}

In this article, a prototype of an expert system was developed dedicated to the monitoring of data network equipment in the context of the remote control of electrical systems. This expert system treats inconsistencies explicitly, that could arise from different measurements related to data network monitoring equipment parameters. In a context where there are several monitored equipment, contradictions are common occurrences.

The expert system developed in this work is based on Fuzzy logic and Paraconsistent Annotated Logic with Annotation of Two Values and it was verified to be capable of analysing uncertainties in the readings of real engineering quantities, and generating the operating conditions for the equipment of data networks.

It is also important to note that the case studied in this article was applied to electrical energy systems. However, the techniques presented using the combination of paraconsistent annotated and Fuzzy sets can be applied to other industrial installations, in which the remote monitoring is carried out through one or more SCADA systems.

Future work should consider the monitoring of a larger set of variables in each expert subsystem than that considered in the present work, allowing the monitoring of more equipment. This would imply on the implementation of additional expert subsystems and also the use of more sources of information for the engineering quantities (processing, memory, data rate, etc). The monitoring of more variables by the expert  subsystems would allow a more complete analysis of each equipment, generating a more accurate classification of the installation operating condition.

As a suggestion for a larger set of monitoring variables, one can, for example, assume for subsystem (A) the monitoring of the data rate and errors on network interfaces on servers; for the subsystem (B), the number of ICMP packets delivered and not delivered on the network interfaces, and use of memory and processing in routers; and, for the subsystem (C), the quantity of TCP messages with re-transmission and discard, and also the use of memory, processing and allocation of files in the ACUs.

\section*{References}


\section*{Attachments}

 \label{my-label-table5}
 {\footnotesize
 \begin{longtable}[h]{ c c c c c}
 
 \caption{Values of the trapezoidal input functions of the Fuzzy procedure of step 2 of the expert system}.\label{long}\\
 
 \textbf{EQUIPMENT}                & \textbf{\shortstack{IN\\INFOR- \\MATION}}             & \textbf{\shortstack{TRAPEZOIDAL \\FUNCTION 1}}              &  \textbf{\shortstack{TRAPEZOIDAL \\FUNCTION 2}}         & \textbf{\shortstack{TRAPEZOIDAL \\FUNCTION 3}}    \\
 (description)                      & (eng                      & (X1,Y1;X2,Y2;                 &  (X1,Y1;X2,Y2;            & (X1,Y1;X2,Y2;        \\
                                    & unit)                     & X3,Y3;X4,Y4)*                 &  X3,Y3;X4,Y4)*            & X3,Y3;X4,Y4)*        \\
 \hline
 \endfirsthead
 
 \multicolumn{5}{c}{Continuation of Table \ref{long}}\\
 \hline
 \textbf{EQUIPMENT}                & \textbf{\shortstack{IN\\INFOR- \\MATION}}             & \textbf{\shortstack{TRAPEZOIDAL \\FUNCTION 1}}              &  \textbf{\shortstack{TRAPEZOIDAL \\FUNCTION 2}}         & \textbf{\shortstack{TRAPEZOIDAL \\FUNCTION 3}}    \\
 (description)                      & (eng                      & (X1,Y1;X2,Y2;                 &  (X1,Y1;X2,Y2;            & (X1,Y1;X2,Y2;        \\
                                    & unit)                     & X3,Y3;X4,Y4)*                 &  X3,Y3;X4,Y4)*            & X3,Y3;X4,Y4)*        \\
 \hline
 \endhead
 
 \hline
 \endfoot
 
 \hline
 \endlastfoot
 
\multirow{3}{*}{SCATEX SERVER}                              &   proc (\%)       &   (0,0;0,1;15,1;50;0) &   (15,0;50,1;60,1;70,0)   &   (60,0;70,1;100,1;100,0) \\
                                                            &   mem (\%)        &   (0,0;0,1;60,1;65,0) &   (60,0;65,1;90,1;95,0)   &   (90,0;95,1;100,1;100,0) \\
                                                            &   file\_al (\%)   &   (0,0;0,1;75,1;85,0) &   (75,0;85,1;90,1;95,0)   &   (90,0;95,1;100,1;100,0) \\
 \hline
 \multirow{3}{*}{\shortstack{UC500/CLP500\\SCADA SERVER}}   &   proc (\%)       &   (0,0;0,1;15,1;50,0) &   (15,0;50,1;60,1;70,0)   &   (60,0;70,1;100,1;100,0) \\
                                                            &   mem (\%)        &   (0,0;0,1;40,1;50,0) &   (40,0;50,1;60,1;65,0)   &   (60,0;65,1;100,1;100,0) \\
                                                            &   file\_al (\%)   &   (0,0;0,1;75,1;85,0) &   (75,0;85,1;90,1;95,0)   &   (0,0;95,1;100,1;100,0)  \\
 \hline
 \multirow{2}{*}{\shortstack{ROUTER 1 –\\HYDROELEC.\\PLANT 1}}    &   rate (kbps)     &   (0,0;0,1;1,1;6.5,0) &   (1,0;65,1;2048,1;2252,0)    &   \shortstack{(2048,0;2252,1;\\4096,1;4096,0)}    \\
                                                            &   error (octets) &   (0,0;0,1;8,1;12,0)  &   (8,0;12,1;12,1;24,0)        &   (12,0;24,1;45,1;45,0)           \\
 \hline
 \multirow{2}{*}{\shortstack{ROUTER 2 –\\HYDROELEC.\\PLANT 1}}    &   rate (kbps)     &   (0,0;0,1;1,1;6.5,0) &   (1,0;65,1;1024,1;1126,0)    &   \shortstack{(2048,0;2252,1;\\4096,1;4096,0)}    \\
                                                            &   error (octets) &   (0,0;0,1;8,1;12,0)  &   (8,0;12,1;12,1;24,0)        &   (12,0;24,1;45,1;45,0)           \\
 \hline
 \multirow{2}{*}{\shortstack{ROUTER 1 –\\SUBSTATION 1}}     &   rate (kbps)     &   (0,0;0,1;1,1;2.5,0) &   (1,0;25,1;2048,1;2252,0)    &   \shortstack{(2048,0;2252,1;\\4096,1;4096,0)}    \\
                                                            &   error (octets) &   (0,0;0,1;3,1;5,0)   &   (3,0;5,1;5,1;10,0)          &   (5,0;10,1;45,1;45,0)            \\
 \hline
 \multirow{2}{*}{\shortstack{ACU 1 –\\HYDROELEC.\\PLANT 1}}       &   rate (msgs)     &   (0,0;0,1;1,1;9,0)   &   (1,0;9,1;2048,1;2252,0) &   \shortstack{(2048,0;2252,1;\\4096,1;4096,0)}    \\
                                                            &   error (msgs)    &   (0,0;0,1;12,1;18,0) &   (12,0;18,1;18,1;34,0)   &   (18,0;34,1;45,1;45,0)               \\
  \hline
 \multirow{2}{*}{\shortstack{ACU 1 –\\SUBSTATION 1}}        &   rate (msgs)     &   (0,0;0,1;1,1;9,0)   &   (1,0;9,1;2048,1;2252,0) &   \shortstack{(2048,0;2252,1;\\4096,1;4096,0)}    \\
                                                            &   error (msgs)    &   (0,0;0,1;12,1;18,0) &   (12,0;18,1;18,1;34,0)   &   (18,0;34,1;45,1;45,0)               \\

 \end{longtable}}
* X1 and Y1 are the coordinates (abscissa and ordered axis) of point 1 of the trapezoidal function, X2 and Y2 respectively of point 2,
X3 and Y3 of point 3 and X4 and Y4 of point 4.

\vspace{9mm}
 \label{my-label-table6}
 {\footnotesize
 \begin{longtable}[h]{ c c c c c}
 \caption{Trapezoidal function values for the Fuzzy procedure of step 2 of the expert system.}\label{long_1}\\
 
 \textbf{\shortstack{EQUIPMENT/ \\ANALYSIS\\ NODE}}                    & \textbf{\shortstack{OUT \\INFOR- \\MATION}}               & \textbf{\shortstack{TRAPEZOIDAL \\FUNCTION 4}}                 &  \textbf{\shortstack{TRAPEZOIDAL \\FUNCTION 5}}         & \textbf{\shortstack{TRAPEZOIDAL \\FUNCTION 6}}    \\
 \shortstack{(description, \\node\\identification}                      & \shortstack{(evidence\\degree}                            & (X1,Y1;X2,Y2;                                                 &  (X1,Y1;X2,Y2;                                         & (X1,Y1;X2,Y2;        \\
                                                                        &                                                           & X3,Y3;X4,Y4)*                                                 &  X3,Y3;X4,Y4)*                                         & X3,Y3;X4,Y4)*        \\
 \hline
 \endfirsthead
 
 \multicolumn{5}{c}{Continuation of Table \ref{long_1}}\\
 \hline
 \textbf{\shortstack{EQUIPMENT/ \\ANALYSIS\\ NODE}}                    & \textbf{\shortstack{OUT \\INFOR- \\MATION}}               & \textbf{\shortstack{TRAPEZOIDAL \\FUNCTION 4}}                 &  \textbf{\shortstack{TRAPEZOIDAL \\FUNCTION 5}}         & \textbf{\shortstack{TRAPEZOIDAL \\FUNCTION 6}}    \\
 \shortstack{(description, \\node\\identification}                      & \shortstack{(evidence\\degree}                            & (X1,Y1;X2,Y2;                                                 &  (X1,Y1;X2,Y2;                                         & (X1,Y1;X2,Y2;        \\
                                                                        &                                                           & X3,Y3;X4,Y4)*                                                 &  X3,Y3;X4,Y4)*                                         & X3,Y3;X4,Y4)*        \\
 \hline
 \endhead
 
 \hline
 \endfoot
 
 \hline
 \endlastfoot
    
 \multirow{2}{*}{\shortstack{SCATEX SERVER \\/NODE 1}}                                              &   fav 1 ($\mu $)              &   (0,0;0,1;0,1;0.5;0) &   (0,0;0.5,1;0.5,1;1,0)   &   (0.5,0;1,1;1,1;1,0) \\
                                                                                                    &   unfav 1 ($\lambda $)        &   (0,0;0,1;0,1;0.5;0) &   (0,0;0.5,1;0.5,1;1,0)   &   (0.5,0;1,1;1,1;1,0) \\
 \hline
 \multirow{2}{*}{\shortstack{SCATEX SERVER \\/NODE \\2}}                                            &   fav 2 ($\mu $)              &   (0,0;0,1;0,1;0.5;0) &   (0,0;0.5,1;0.5,1;1,0)   &   (0.5,0;1,1;1,1;1,0) \\
                                                                                                    &   unfav 2 ($\lambda $)        &   (0,0;0,1;0,1;0.5;0) &   (0,0;0.5,1;0.5,1;1,0)   &   (0.5,0;1,1;1,1;1,0) \\
 
 \hline
 \multirow{2}{*}{\shortstack{UC500/CLP500\\SCADA SERVER\\/NODE 1}}                                  &   fav 1 ($\mu $)              &   (0,0;0,1;0,1;0.5;0) &   (0,0;0.5,1;0.5,1;1,0)   &   (0.5,0;1,1;1,1;1,0) \\
                                                                                                    &   unfav 1 ($\lambda $)        &   (0,0;0,1;0,1;0.5;0) &   (0,0;0.5,1;0.5,1;1,0)   &   (0.5,0;1,1;1,1;1,0) \\
 \hline
 \multirow{2}{*}{\shortstack{UC500/CLP500\\SCADA SERVER\\/NODE 2}}                                  &   fav 2 ($\mu $)              &   (0,0;0,1;0,1;0.5;0) &   (0,0;0.5,1;0.5,1;1,0)   &   (0.5,0;1,1;1,1;1,0) \\
                                                                                                    &   unfav 2 ($\lambda $)        &   (0,0;0,1;0,1;0.5;0) &   (0,0;0.5,1;0.5,1;1,0)   &   (0.5,0;1,1;1,1;1,0) \\

 \hline
 \multirow{2}{*}{\shortstack{ROUTER 1 –\\HYDROELEC.\\PLANT 1/NODE 1}}                                   &   fav 1 ($\mu $)              &   (0,0;0,1;0,1;0.5;0) &   (0,0;0.5,1;0.5,1;1,0)   &   (0.5,0;1,1;1,1;1,0) \\
                                                                                                    &   unfav 1 ($\lambda $)        &   (0,0;0,1;0,1;0.5;0) &   (0,0;0.5,1;0.5,1;1,0)   &   (0.5,0;1,1;1,1;1,0) \\
 \hline
 \multirow{2}{*}{\shortstack{ROUTER 2 –\\HYDROELEC.\\PLANT 1/NODE 1}}                                   &   fav 1 ($\mu $)              &   (0,0;0,1;0,1;0.5;0) &   (0,0;0.5,1;0.5,1;1,0)   &   (0.5,0;1,1;1,1;1,0) \\
                                                                                                    &   unfav 1 ($\lambda $)        &   (0,0;0,1;0,1;0.5;0) &   (0,0;0.5,1;0.5,1;1,0)   &   (0.5,0;1,1;1,1;1,0) \\
 \hline
 \multirow{2}{*}{\shortstack{ROUTER 1 –\\SUBSTATION 1\\/NODE 1}}                                    &   fav 1 ($\mu $)              &   (0,0;0,1;0,1;0.5;0) &   (0,0;0.5,1;0.5,1;1,0)   &   (0.5,0;1,1;1,1;1,0) \\
                                                                                                    &   unfav 1 ($\lambda $)        &   (0,0;0,1;0,1;0.5;0) &   (0,0;0.5,1;0.5,1;1,0)   &   (0.5,0;1,1;1,1;1,0) \\
 \hline
 \multirow{2}{*}{\shortstack{ACU 1 –\\HYDROELEC.\\PLANT 1/NODE 1}}                                      &   fav 1 ($\mu $)              &   (0,0;0,1;0,1;0.5;0) &   (0,0;0.5,1;0.5,1;1,0)   &   (0.5,0;1,1;1,1;1,0) \\
                                                                                                    &   unfav 1 ($\lambda $)        &   (0,0;0,1;0,1;0.5;0) &   (0,0;0.5,1;0.5,1;1,0)   &   (0.5,0;1,1;1,1;1,0) \\
 \hline
 \multirow{2}{*}{\shortstack{ACU 1 –\\SUBSTATION 1\\/NODE 1}}                                       &   fav 1 ($\mu $)              &   (0,0;0,1;0,1;0.5;0) &   (0,0;0.5,1;0.5,1;1,0)   &   (0.5,0;1,1;1,1;1,0) \\
                                                                                                    &   unfav 1 ($\lambda $)        &   (0,0;0,1;0,1;0.5;0) &   (0,0;0.5,1;0.5,1;1,0)   &   (0.5,0;1,1;1,1;1,0) \\

 \end{longtable}}
* same comment from Table 1 in the Attachments.

\vspace{100mm}
 \label{my-label-table7}
 {\footnotesize
 \begin{longtable}[c]{ c c c}
 \caption{Association rules for trapezoidal functions for entering and leaving the Fuzzy procedure step 2 of the expert system.}\label{long_2}\\
 
 \textbf{EQUIPMENT}                & \textbf{RULES}                                                                & \textbf{\shortstack{ACTING \\FUNCTIONS}}  \\
 (description)                      & \shortstack{(association rules of\\input information)}                        & \shortstack{(degrees evidence)}           \\
                                    &                                                                               &                                           \\
 \hline
 \endfirsthead
 
 \multicolumn{3}{c}{Continuation of Table \ref{long_2}}\\
 \hline
 \textbf{EQUIPMENT}                & \textbf{RULES}                                                                & \textbf{\shortstack{ACTING \\FUNCTIONS}}  \\
 (description)                      & \shortstack{(association rules of\\input information)}                        & \shortstack{(degrees evidence)}           \\
                                    &                                                                               &                                           \\
 \hline
 \endhead
 
 \hline
 \endfoot
 
 \hline
 \endlastfoot
 
 \multirow{6}{*}{\shortstack{SERVERS}}                                                                  &   (proc function 1 \textbf{AND} mem function 1) \textbf{OR}                                                   &   \multirow{3}{*}{\shortstack{(function 4\\of fav 1)\\ \textbf{AND}
(function 6\\of unfav 1)}}  \\
                                                                                                        &   (proc function 1 \textbf{AND} mem function 2) \textbf{OR}                                                   &   \\  
                                                                                                        &   (proc function 2 \textbf{AND} mem function 1)                                                               &   \\  \cline{2-3}
                                                                                                        &   (proc function 2 \textbf{AND} mem function 2)                                                               &   \shortstack{(function 5\\of fav 1)\\ \textbf{AND}
(function 6\\of unfav 1)}   \\  \cline{2-3}
                                                                                                        &   \shortstack{(other combinations between processing \\ and memory functions)}                                 &   \shortstack{(function 6\\of fav 1)\\ \textbf{AND}
(function 4\\of unfav 1)}   \\  \cline{2-3} \vspace{5pt}
                                                                                                        &   \multicolumn{2}{c}{}{\textbf{the same rules above, but considering processing and file allocation}} \\

 \hline
 \multirow{5}{*}{\shortstack{ROUTERS}}                                                                  &   (rate function 2 \textbf{AND} error function 1) \textbf{OR}                                                 &   \multirow{3}{*}{\shortstack{(function 4\\of fav 1)\\ \textbf{AND}
(function 6\\of unfav 1)}}  \\
                                                                                                        &   (rate function 3 \textbf{AND} error function 1) \textbf{OR}                                                 &   \\
                                                                                                        &   (rate function 3 \textbf{AND} error function 2)                                                             &   \\  \cline{2-3}
                                                                                                        &   (rate function 2 \textbf{AND} error function 2)                                                             &   \shortstack{(function 5\\of fav 1)\\ \textbf{AND}
(function 5\\of unfav 1)}   \\  \cline{2-3}
                                                                                                        &   \shortstack{(other combinations between the rate \\ and error functions)}                                    &   \shortstack{(function 6\\of fav 1)\\ \textbf{AND}
(function 4\\of unfav 1)}   \\

 \hline
 \multirow{6}{*}{\shortstack{ACUs}}                                                                     &   (rate function 2 \textbf{AND} error function 1) \textbf{OR}                                                 &   \multirow{3}{*}{\shortstack{(function 4\\of fav 1)\\ \textbf{AND}
(function 6\\of unfav 1)}}  \\
                                                                                                        &   (rate function 3 \textbf{AND} error function 1)                                                             &   \\
                                                                                                        &                                                                                                               &   \\  \cline{2-3}
                                                                                                        &   (rate function 2 \textbf{AND} error function 2)                                                             &   \multirow{2}{*}{\shortstack{(function 5\\of fav 1)\\ \textbf{AND}
(function 5\\of unfav 1)}}  \\
                                                                                                        &   (rate function 3 \textbf{AND} error function 2)                                                             &   \\
                                                                                                        &                                                                                                               &   \\  \cline{2-3}
                                                                                                        &   \shortstack{(other combinations between the rate \\ and error functions)}                                    &   \shortstack{(function 6\\of fav 1)\\ \textbf{AND}
(function 4\\of unfav 1)}   \\
 \end{longtable}}

\end{document}